\documentclass[12pt,twocolumn,twoside]{IEEEtran}

\pdfoutput=1

\usepackage{grffile} 
\usepackage{graphicx}
\usepackage{xcolor}
\usepackage[cmex10]{amsmath}
\usepackage{amssymb}
\usepackage{algorithm,algorithmic,alltt}
\usepackage{array}
\usepackage{booktabs}
\usepackage{lineno}
\usepackage{multirow}
\usepackage[caption=false,font=footnotesize,subrefformat=parens,labelformat=parens]{subfig}
\usepackage{fixltx2e}
\usepackage{stfloats}
\usepackage[abs]{overpic}
\usepackage{epstopdf}
\usepackage{hyperref}

\DeclareMathOperator*{\argmin}{argmin}

\definecolor{ColorCloseup0}{rgb}{1, 0, 0}
\definecolor{ColorCloseup1}{rgb}{0, 0.5, 1}
\definecolor{ColorCloseup2}{rgb}{0, 0.5, 0.75}
\definecolor{ColorCloseup3}{rgb}{0.5, 0.5, 1}
\definecolor{ColorCloseup4}{rgb}{0, 0, 0.63}

\def\swidthone{0.97\linewidth}
\def\swidthtwo{0.48\linewidth}
\def\swidththree{0.3\linewidth}
\def\swidthfour{0.23\linewidth}
\def\swidthfive{0.18\linewidth}

\begin{document}
\title{Robust Piecewise-Constant Smoothing: $M$-Smoother Revisited
\thanks{L. Bao is a Ph.D. student with the Department of Computer Science at the City University of Hong Kong, Hong Kong (e-mail: linchaobao@gmail.com).}
\thanks{Q. Yang is an Assistant Professor with the Department of Computer Science at the City University of Hong Kong, Hong Kong (e-mail: qiyang@cityu.edu.hk).}
\thanks{The Matlab code can be found in the author's homepage: \url{https://sites.google.com/site/linchaobao/}.
}
}
\author{
        Linchao~Bao and Qingxiong~Yang \\
{\small\texttt{\url{https://sites.google.com/site/linchaobao/}}}
}
\maketitle

\begin{abstract}
A robust estimator, namely $M$-smoother, for piecewise-constant smoothing is revisited in this paper. Starting from its generalized formulation, we propose a numerical scheme/framework for solving it via a series of weighted-average filtering (\emph{e.g.}, box filtering, Gaussian filtering, bilateral filtering, and guided filtering). Because of the equivalence between $M$-smoother and local-histogram-based filters (such as median filter and mode filter), the proposed framework enables fast approximation of histogram filters via a number of box filtering or Gaussian filtering. In addition, high-quality piecewise-constant smoothing can be achieved via a number of bilateral filtering or guided filtering integrated in the proposed framework. Experiments on depth map denoising show the effectiveness of our framework.
\end{abstract}

\begin{keywords}
Piecewise-constant smoothing, $M$-smoother, Edge-preserving filter, Bilateral filter, Guided filter


\end{keywords}

\section{Introduction}
Piecewise-constant smoothing serves as a fundamental tool in many image processing and low-level vision tasks. Originating from different background, a wide range of techniques are proposed to solve the problem, including anisotropic diffusion \cite{perona1990scale}, bilateral filtering \cite{tomasi1998bilateral}, robust estimation \cite{chu1998edge}, etc. Their relations have been widely discussed in the literature \cite{black1998robust,winkler1999noise,barash2002fundamental,elad2002origin,mrazek2006robust} to benefit each other. For example, anisotropic diffusion is improved by exploiting new ``edge-stopping'' functions based on the well-studied influence functions from robust statistics \cite{black1998robust}. We in this paper focus on exploring the reciprocity between robust estimation and weighted-average filters.

Specifically, We show that a robust estimator -- the $M$-smoother \cite{chu1998edge} -- can be reformulated as a series of weighted-average filtering followed by a winner-take-all operation. As a result, it can be much more efficiently approximated utilizing existing fast filtering algorithms. More importantly, existing weighted-average based edge-preserving filters, such as bilateral filter, guided filter \cite{he2010guided}, and cross-multi-point filter \cite{lu2012cross}, can be largely ``enhanced'' to achieve high-quality piecewise-constant smoothing when integrated into the framework with well-studied robust loss functions.

The paper is organized as follows. In the next section we briefly review the preliminaries of weighted-average filters, local-histogram-based filters and $M$-smoother (note that local-histogram-based filters are closely related with $M$-smoother \cite{mrazek2006robust}). We present our framework in Sec. \ref{sec:approach}. Sec. \ref{sec:analysis} provides more discussion by exploiting the specific forms of the proposed framework. Finally, Sec. \ref{sec:conclusion} concludes the paper.

\section{Preliminaries and Notions}

\subsection{Weighted-average Filters}
Throughout this paper, we use $\mathcal{F}(\cdot)$ to denote a filter operation performed on an image, that is, assuming $I$ to be an input image and $J$ the filtered image,
\begin{equation}
J = \mathcal{F}(I).
\end{equation}
A wide range of weighted-average filters can be expressed as, denoting $I_\mathbf{p}$ the color/intensity value at pixel $\mathbf{p}$ and $J_\mathbf{p}$ the filtered value at pixel $\mathbf{p}$,
\begin{equation}
    J_\mathbf{p} = \sum\limits_{\mathbf{q}\in\Omega_\mathbf{p}} w_{\mathbf{pq}}^{(T)} \cdot I_\mathbf{q},
    \label{eq.generalwaf}
\end{equation}
where $\Omega_\mathbf{p}$ is the neighborhood of pixel $\mathbf{p}$ and the weighting function $w_{\mathbf{pq}}^{(T)}$ may or may not depend on a guidance image $T$. For example, when the weighting function is a normalized Gaussian function depending on spatial distance between $\mathbf{p}$ and $\mathbf{q}$, the formulation is a Gaussian filter. When the color/intensity value of pixel $\mathbf{p}$ and $\mathbf{q}$ in guidance image $T$ is taken into account, the formulation becomes the bilateral filter\footnote{Bilateral filter is also referred to as $\sigma$-filter in \cite{chu1998edge} or nonlinear Gaussian filter in \cite{winkler1999noise}.} \cite{tomasi1998bilateral}. See Table \ref{tab:wafall} for the weighted-average filters considered in this paper. Note that although the weighting function of the guided filter \cite{he2010guided} seems a little complicated, it actually has a rather efficient algorithm (the complexity is about 6 times as much as that of a box filtering for single channel image).

\begin{table}[!ht]
 \footnotesize
 \center
 \caption{Weighted-average filters\label{tab:wafall}}
 \begin{tabular}{ll}
  \toprule
  Weighting function & Filter \\
  \midrule
  $w_{\mathbf{pq}}^{(T)}=\frac{1}{W_\mathbf{p}}$   & Box (BF)$^\dag$ \\
  $w_{\mathbf{pq}}^{(T)}=\frac{1}{W_\mathbf{p}}G_\sigma(\|\mathbf{p}-\mathbf{q}\|)$   & Gaussian (GF)$^\dag$ \\
  $w_{\mathbf{pq}}^{(T)}=\frac{1}{W_\mathbf{p}}G_{\sigma_s}(\|\mathbf{p}-\mathbf{q}\|)G_{\sigma_r}(|T_\mathbf{p}-T_\mathbf{q}|)$   & Bilateral (BLF)$^\dag$ \\
  $w_{\mathbf{pq}}^{(T)}=\frac{1}{|\omega|^2} \sum\limits_{k:(\mathbf{p},\mathbf{q}) \in \omega_k} (1 + \frac{(T_\mathbf{p} - \mu_k)(T_\mathbf{q} - \mu_k)}{\sigma_k^2 + \epsilon})$   & Guided (GDF)$^\ddag$ \\
  \bottomrule
  \multicolumn{2}{l}{\multirow{1}{0.98\linewidth}{\hspace{-6pt}\scriptsize{$\dag$: Notion $W_\mathbf{p}$ is normalization factor (summing all weights for pixel $\mathbf{p}$), $G(\cdot)$ is the Gaussian function.}}} \\ & \\
  \multicolumn{2}{l}{\multirow{2}{0.98\linewidth}{\hspace{-6pt}\scriptsize{$\ddag$: Notion $\omega_k$ denotes each window (with radius $r$ and $|\omega|$ pixels) that covers both pixel $\mathbf{p}$ and $\mathbf{q}$, whose mean and variance are $\mu_k$ and $\sigma_k$, respectively. $\epsilon$ is a parameter.}}} \\ \\
 \end{tabular}
\end{table}

The parameters for the four filters are as follows: box filter is controlled by the radius $r$ of the box window; Gaussian filter is controlled by the parameter $\sigma$ in Gaussian function; bilateral filter is controlled by parameter $\sigma_s$ of spatial kernel and parameter $\sigma_r$ of range kernel; guided filter is controlled by $r$ and $\epsilon$. The parameter ``correspondence'' between bilateral filter and guided filter is suggested in \cite{he2010guided}: $\sigma_s \leftrightarrow r$ and $\sigma_r^2 \leftrightarrow \epsilon$. For a unified discussion, we further extend the ``correspondence'' to box filter and Gaussian filter: we also use $\sigma_s$ to denote the parameter $\sigma$ in Gaussian filter and control the parameter $r$ in box filter\footnote{To achieve the same amount of smoothing as Gaussian filter, the $r$ in box filter is calculated by $r=\lfloor \sqrt{2} \sigma_s \rfloor$ empirically in our experiments \cite{he2013guidedpami} (it can be verified by experimental comparison using PSNR between box filtered image and Gaussian filtered image).}. In this way, we can use the two parameters $\sigma_s$ and $\sigma_r$ to discuss all the filters, following the convention of the bilateral filter \cite{paris2009bilateralbook} ($\sigma_s$ is measured by pixel number and $\sigma_r$ is a real number between $0$ and $1$ indicating a fraction of the whole intensity range $\mathcal{R}$).

\subsection{Local-histogram-based Filters}
A family of local-histogram-based filters \cite{van2001local,kass2010smoothed} (e.g., median filter and mode filters) are to replace the color/intensity of each pixel with the color/intensity of neighboring majority pixels (e.g., using some certain robust statistics drawn out from local histogram). For example, median filtering is to replace each pixel value with the median of neighboring pixel values (if Gaussian weighted neighborhood is used, it is called \emph{isotropic} median filtering \cite{kass2010smoothed}). The \emph{closest-mode} filtering \cite{kass2010smoothed} (also referred to as \emph{local mode} filtering in \cite{van2001local}) replaces each pixel with its closest mode, and the \emph{dominant-mode} filtering \cite{kass2010smoothed} (similar to the \emph{global mode} filtering in \cite{van2001local}) instead uses the mode having the largest population. Although such kind of filters can smooth out high contrast, fine-scale details, they often face a problem of serious deviation from the original edges (especially at corners), since local histogram completely ignores image geometric structures.

For convenience, in the rest of this paper, we use the term ``local mode filter'' and ``global mode filter'' in \cite{van2001local} to refer to filters whose local histograms are constructed within \emph{hard} spatial windows, and use the term ``closest-mode filter'' and ``dominant-mode filter'' in \cite{kass2010smoothed} to refer to filters whose local histograms are constructed within Gaussian weighted \emph{soft} spatial windows.

\subsection{$M$-Smoother}\label{sec:relatedworkrobust}
From the statistical point of view, the simplest, non-robust estimation of the underlying image signal contaminated by zero-mean Gaussian noise is to estimate the intensity value of each pixel by minimizing the sum of squared residual errors ($L_2$ norm) within local window
\begin{equation}
    J_\mathbf{p} = \argmin_{\theta}{\sum\limits_{\mathbf{q}\in\Omega_\mathbf{p}} (\theta-I_\mathbf{q})^2}.
    \label{eq.l2estimate}
\end{equation}
Solving the optimization problem yields exactly the box filter. The problem of such least-squares estimation is that it is very sensitive to outliers and thus pixels at the two sides of an edge will affect each other. As a result, edges will get blurred.

In order to increase \emph{robustness} and reject outliers, the quadratic function in the above formulation need to be replaced with more tolerant function that gives less penalty to outliers. For example, using the absolute error function instead of the quadratic function, we obtain ($L_1$ norm)
\begin{equation}
    J_\mathbf{p} = \argmin_\theta{\sum\limits_{\mathbf{q}\in\Omega_\mathbf{p}} |\theta-I_\mathbf{q}|},
    \label{eq.l1estimate}
\end{equation}
whose solution yields exactly the median filter \cite{winkler1999noise}.

More generally, the above formulation can be extended to the $M$-\emph{smoother} \cite{chu1998edge}\footnote{The $M$-smoother in \cite{chu1998edge} means to find \emph{local} minima that is closest to the original input pixel value when minimizing the objective function. We in this paper do not mean to find local minima, but rather to find the \emph{global} minima of the objective function.}  (originated from the $M$\emph{-estimator} in robust statistics \cite{huber1981robust,hampel1986robust})
\begin{equation}
    J_\mathbf{p} = \argmin_\theta{\sum\limits_{\mathbf{q}\in\Omega_\mathbf{p}} \rho(\theta-I_\mathbf{q})},
    \label{eq.localmsmoother}
\end{equation}
where $\rho(\cdot)$ is a \emph{loss function} (also referred to as $\rho$\emph{-function} or \emph{error norm}). For robustness, the loss function should not grow too rapidly, hence lessening the influence of outliers. Its derivative function is a good tool for studying the influence of outliers, which is often referred to as \emph{influence function} or $\psi$\emph{-function} in robust statistics \cite{huber1981robust,hampel1986robust}. In order to preserve sharp edges in images, a \emph{redescending} influence function ($|\psi(x)| \rightarrow 0$ as $|x| \rightarrow \infty$) is often preferred \cite{chu1998edge,black1998robust} (we hereafter refer to the loss function whose influence function is redescending as \emph{redescending-influence loss function}). Table \ref{tab.lossfuncshow} shows several pairs of the loss function and the corresponding influence function. More robust loss functions can be found in \cite{black1996unification}. Note the parameter $\sigma$ in Table \ref{tab.lossfuncshow} is used to control the influence scale \cite{black1998robust}. In this paper, we associate the scale parameter $\sigma$ in loss function with the range parameter $\sigma_r$ in weighted-average filters (let $\sigma$ = $\sigma_r \cdot \mathcal{R}$), as basically both of them are to control the ``robustness'' \cite{durand2002fast}.

\begin{table}[!t]
    \footnotesize
    \center
	\caption{Loss function and influence function}
    \begin{tabular}{>{\scriptsize}rc|c}
        \toprule
        \multicolumn{2}{c|}{\footnotesize{Loss function$^\dag$}} & \footnotesize{Influence function$^\dag$} \\
        \hline
        &&\\
        $L_1$ norm:  & \multirow{2}{\swidthfour}{\parbox[c]{\swidthone}{\includegraphics[width=\swidthone]{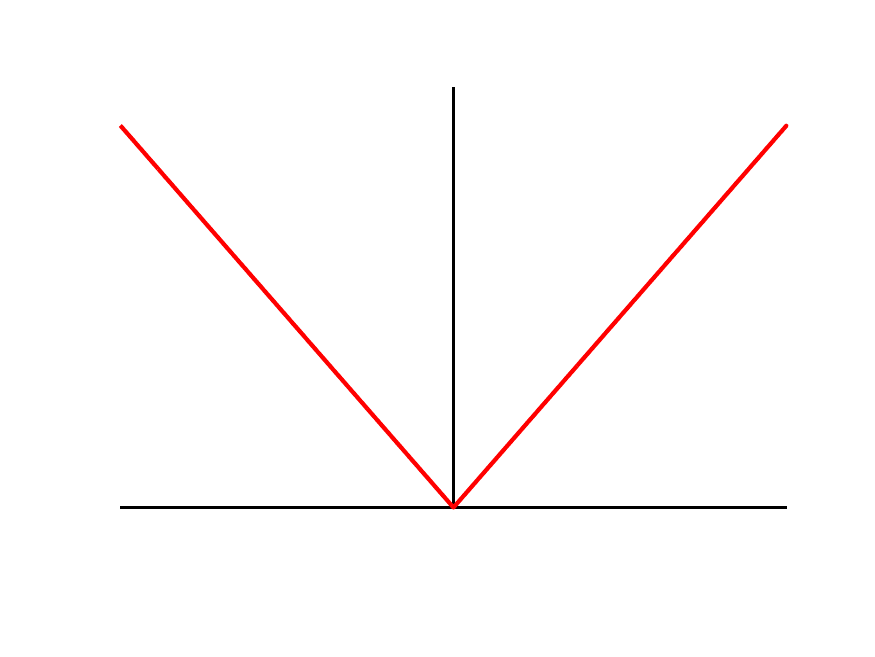}}} & \multirow{2}{\swidthfour}{\parbox[c]{\swidthone}{\includegraphics[width=\swidthone]{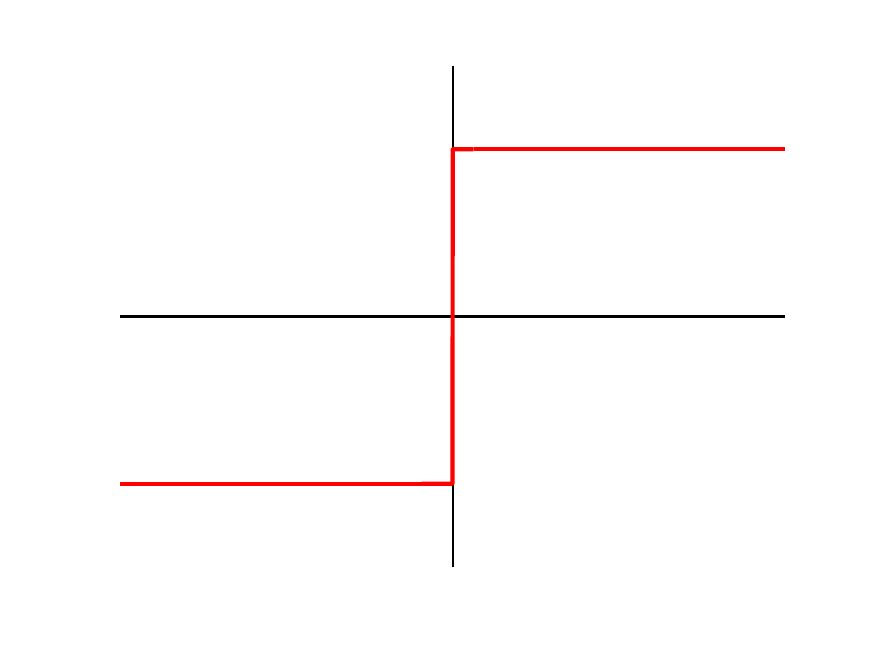}}} \\
        $\rho(x)=|x|$ & & \\ & & \\ & & \\
        Truncated $L_1$ norm: & \multirow{2}{\swidthfour}{\parbox[c]{\swidthone}{\includegraphics[width=\swidthone]{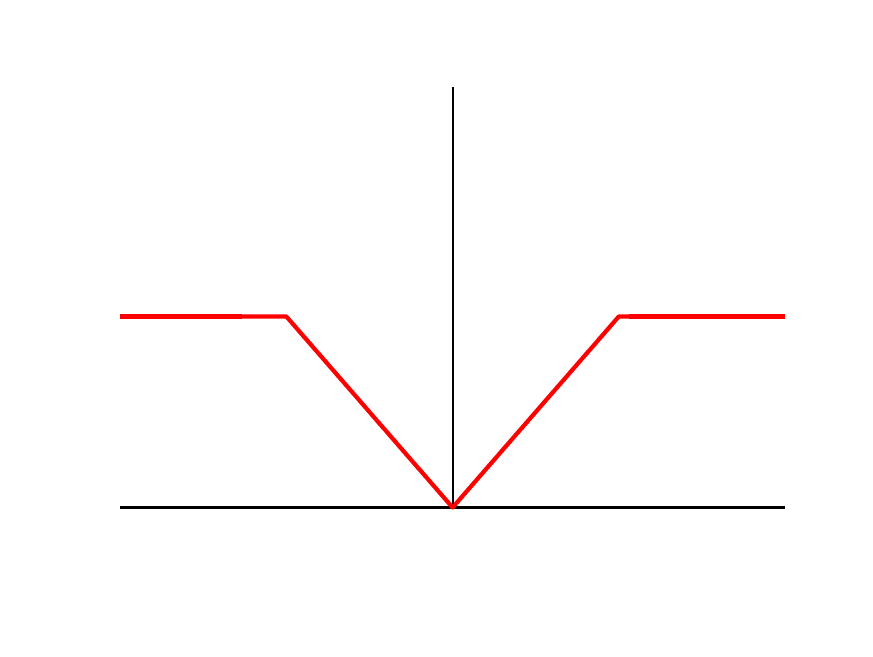}}} & \multirow{2}{\swidthfour}{\parbox[c]{\swidthone}{\includegraphics[width=\swidthone]{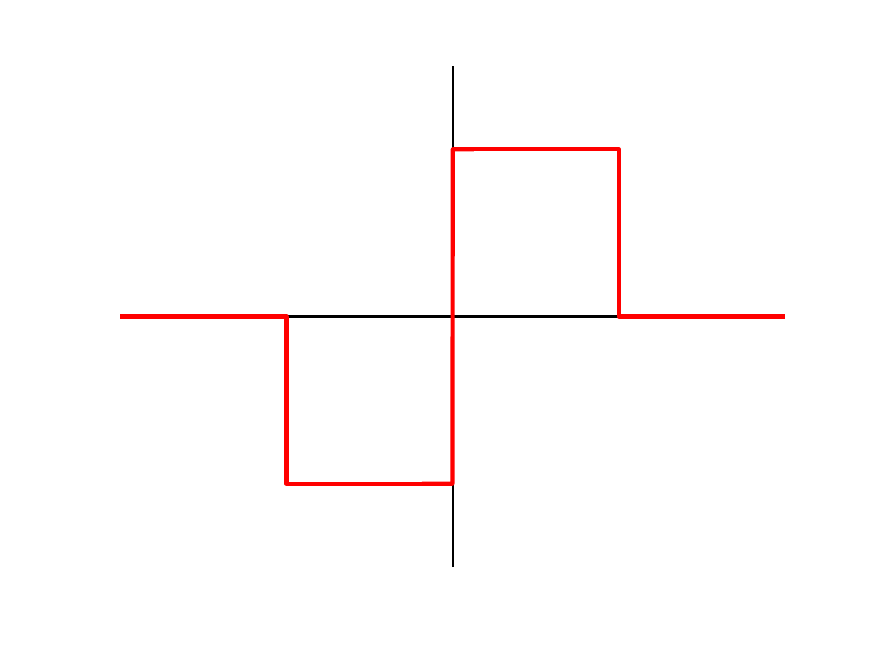}}} \\
        $\rho(x,\sigma) = \left\{ \begin{matrix} |x| \text{,~~if~} |x| \leq \sigma \\ \sigma \text{,~~otherwise~} \end{matrix} \right.$ & & \\ & & \\
        Negative Gauss: & \multirow{2}{\swidthfour}{\parbox[c]{\swidthone}{\includegraphics[width=\swidthone]{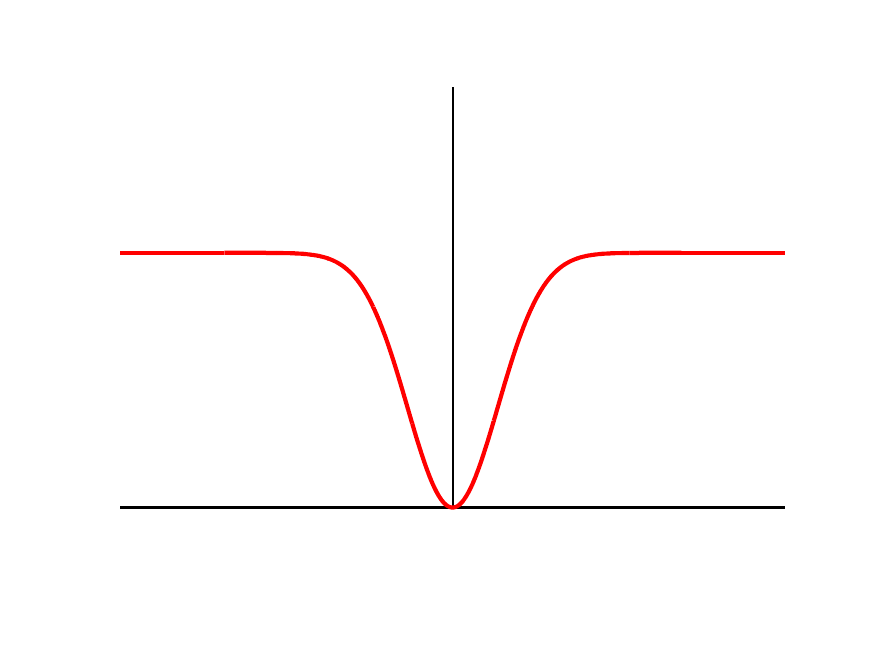}}} & \multirow{2}{\swidthfour}{\parbox[c]{\swidthone}{\includegraphics[width=\swidthone]{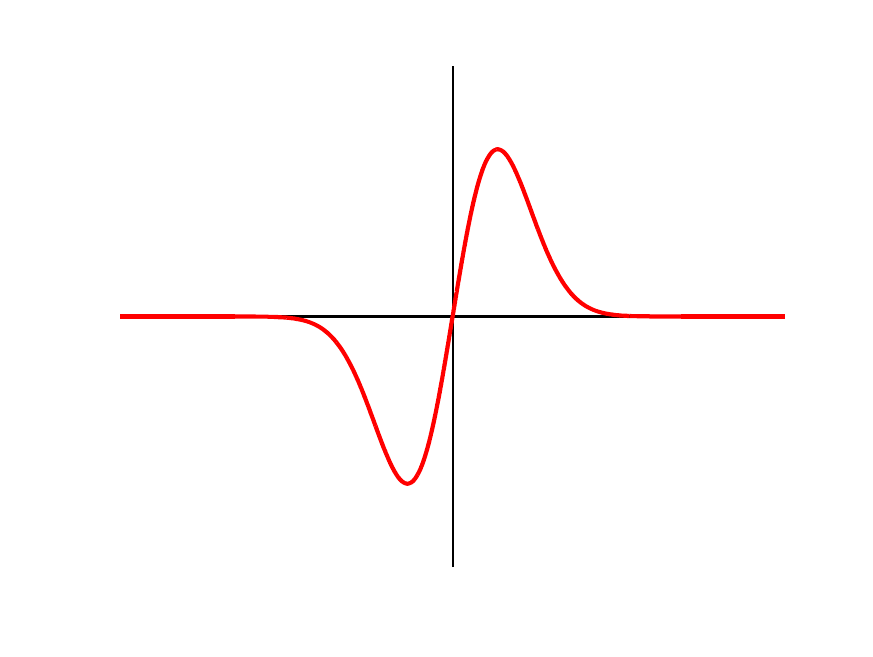}}} \\
        $\rho(x,\sigma) = 1 - e^{-\frac{x^2}{(0.64 \sigma)^2}}$ & & \\ & & \\
        Tukey's biweight: $\rho(x,\sigma) = $  & \multirow{2}{\swidthfour}{\parbox[c]{\swidthone}{\includegraphics[width=\swidthone]{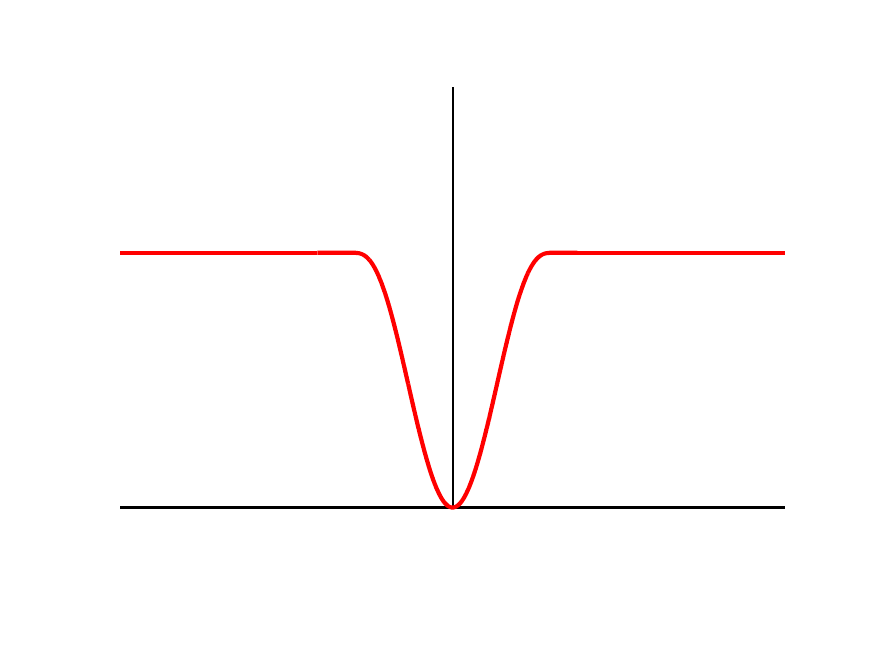}}} & \multirow{2}{\swidthfour}{\parbox[c]{\swidthone}{\includegraphics[width=\swidthone]{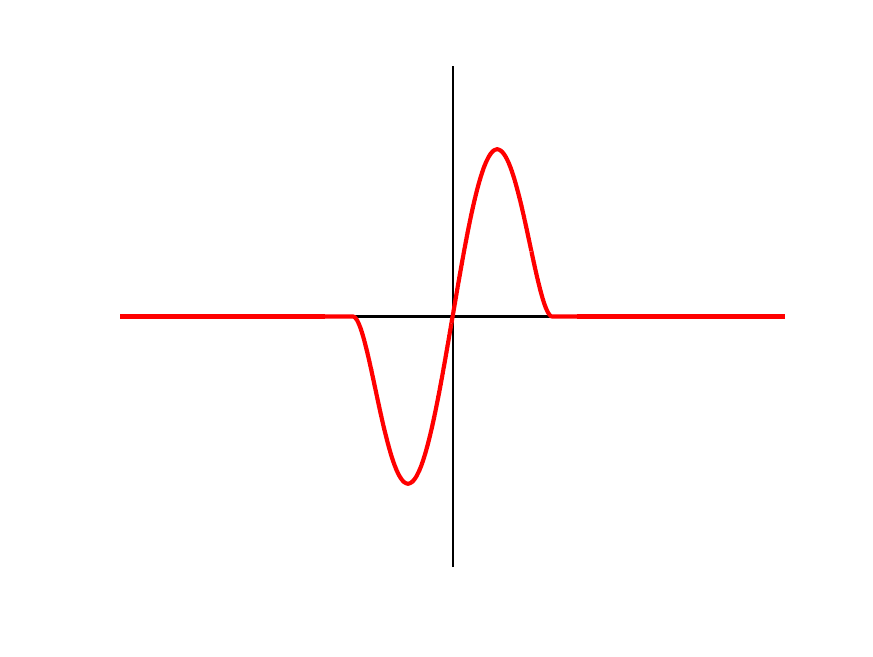}}} \\
        $\left\{ \begin{matrix} \frac{x^2}{\sigma^2} - \frac{x^4}{\sigma^4} + \frac{x^6}{3 \sigma^6} \text{,~if~} |x| \leq \sigma \\ \frac{1}{3} \text{,~~otherwise~} \end{matrix} \right.$ & & \\ & & \\
        Geman-Reynolds:  & \multirow{2}{\swidthfour}{\parbox[c]{\swidthone}{\includegraphics[width=\swidthone]{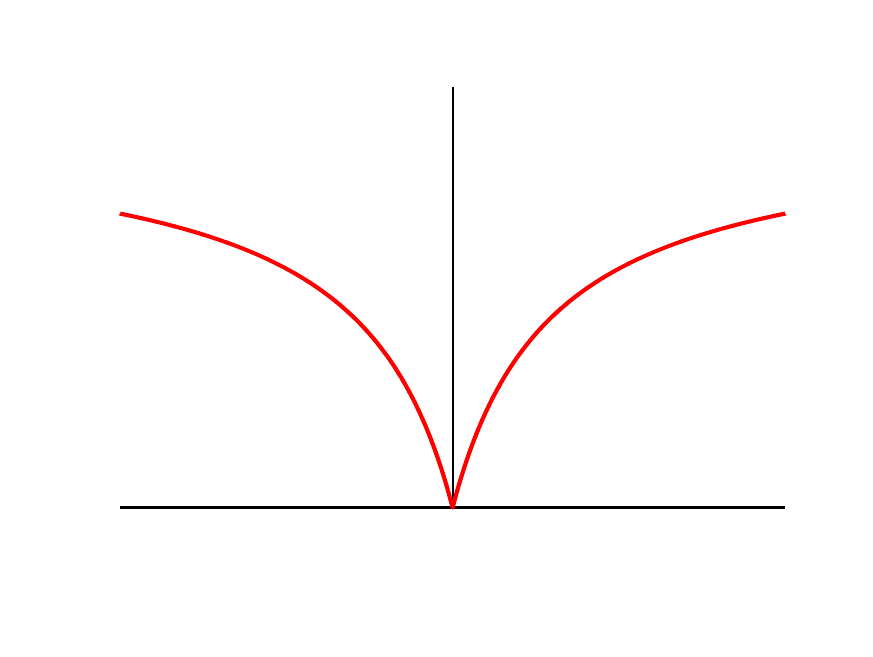}}} & \multirow{2}{\swidthfour}{\parbox[c]{\swidthone}{\includegraphics[width=\swidthone]{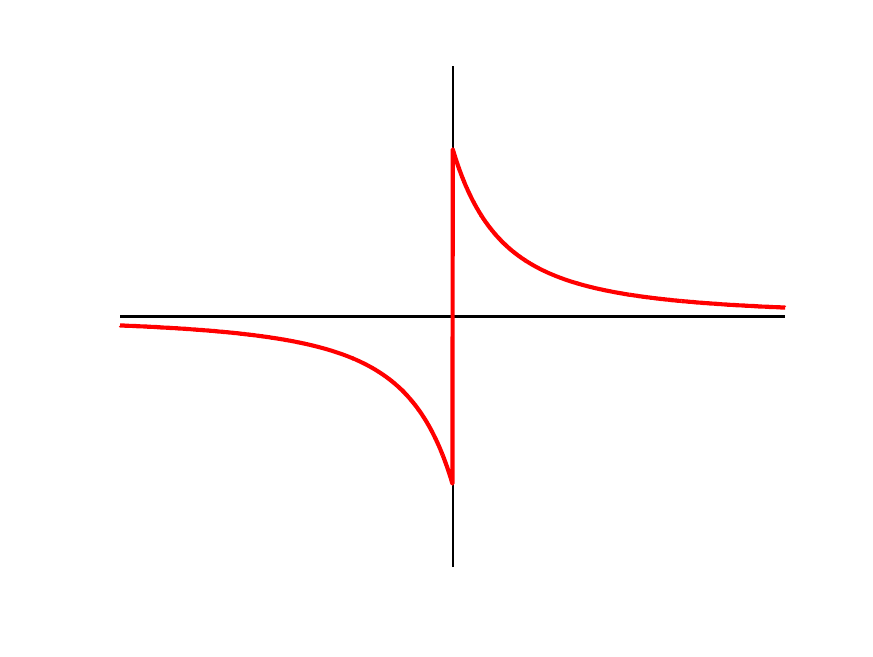}}} \\
        $\rho(x,\sigma) = \frac{-\sigma}{\sigma + |x|}$ & & \\ & & \\
        \bottomrule
        \multicolumn{3}{l}{\multirow{1}{0.98\linewidth}{\hspace{-6pt}\scriptsize{$\dag$: Influence function is the derivative of loss function. Note that the last four loss functions are \emph{redescending-influence loss functions} (RILF).}}} \\
    \end{tabular}
    \label{tab.lossfuncshow}
\end{table}

Additionally, the $M$-smoother can be extended \cite{chu1998edge} to take into account spatial weights in local window. The formulation further becomes
\begin{equation}
    J_\mathbf{p} = \argmin_\theta{\sum\limits_{\mathbf{q}\in\Omega_\mathbf{p}} \rho(\theta-I_\mathbf{q}) \cdot G(\|\mathbf{p}-\mathbf{q}\|)},
    \label{eq.localmsmootherspatial}
\end{equation}
where $G(\cdot)$ is a Gaussian weighting function.

The strategy of employing a robust loss function to reject outliers is analogous to the range weighting function in bilateral filter \cite{winkler1999noise,durand2002fast}. Durand and Dorsey \cite{durand2002fast} noticed this and propose to improve bilateral filter by replacing the Gaussian weighting function with a superior function -- \emph{Tukey's biweight} -- whose influence function is more conform to the redescending principle. Note that although the correspondence between $M$-smoother and bilateral filter can be established in this way, their output are usually much different from each other, since the bilateral filter can only be viewed as one step towards finding \emph{local} minima of the objective function of $M$-smoother using iterative methods \cite{winkler1999noise,mrazek2006robust}. Simply increasing iterations of bilateral filtering still cannot correctly lead to the output of $M$-smoother, as the image is transformed after each step \cite{winkler1999noise}.

Actually, the global minima of $M$-smoother with a negative Gaussian loss function is similar to that of a global mode filter \cite{van2001local} or dominant-mode filter, and the local minima of $M$-smoother corresponds to the local mode filter or closest-mode filter \cite{kass2010smoothed}. Please refer to \cite{mrazek2006robust} for a detailed discussion.

\section{Our Filtering Framework}\label{sec:approach}
In this section, we generalize the $M$-smoother and propose a numerical scheme to solve it via a series of weighted-average filtering. The numerical scheme is indeed our proposed filtering framework, the specific forms of which will be exploited in next section.

\subsection{Generalized $M$-Smoother}
We extend the weighting function of $M$-smoother in Eq. \eqref{eq.localmsmootherspatial} into a more general form:
\begin{equation}
    J_\mathbf{p} = \argmin_\theta{\sum\limits_{\mathbf{q}\in\Omega_\mathbf{p}} \rho(\theta-I_\mathbf{q}) \cdot w_{\mathbf{pq}}^{(T)}},
    \label{eq.extendedsmoother}
\end{equation}
where the $w_{\mathbf{pq}}^{(T)}$ is the weight of pixel $\mathbf{q}$ contributing to $\mathbf{p}$ and $T$ can be either a guidance image (see Table \ref{tab:wafall}) or the input image $I$ itself. Note that our formulation actually combines the \emph{implicit} piecewise-constant model (from robust estimator) and the \emph{explicit} weighting scheme (from weighted-average filters). If the weighting function is the edge-preserving weighting from bilateral filter or guided filter, our formulation can achieve better edge preservation in piecewise-constant smoothing than traditional $M$-smoother due to the explicit weighting scheme (see Sec. \ref{sec:specificblfgf} for details).

\subsection{Solve via a Filtering Framework}
Let $\Theta$ denote the set of all possible output pixel values of the smoother, for a given $\theta \in \Theta$, we define $D(\theta)$ as a \emph{cost image} where each pixel $\mathbf{p}$ is computed from input image $I$ as follows
\begin{equation}
    [D(\theta)]_\mathbf{p} = \rho(\theta-I_\mathbf{p}).
    \label{eq.costimagedef}
\end{equation}
Then we can reformulate Eq. \eqref{eq.extendedsmoother} into
\begin{equation}
    J_\mathbf{p} = \argmin_{\theta \in \Theta}{\left[\mathcal{F} \left( D(\theta) \right) \right]_\mathbf{p}},
    \label{eq.ourframework}
\end{equation}
where $\mathcal{F}(\cdot)$ is a weighted-average filter (see Eq. \eqref{eq.generalwaf}). Assuming set $\Theta$ is finite, the formulation essentially means the filtering on all possible cost images $\{ D(\theta) \mid \theta \in \Theta \}$ is first performed and then an $\argmin$ operation at each pixel $\mathbf{p}$ is individually applied according to the filtered results at $\mathbf{p}$. The process is known as \emph{cost-volume filtering framework} \cite{rhemann2011fast} in the context of discrete labeling problem like stereo matching.


The key insight of the above reformulation is that the filtering on cost image allows us to apply fast filtering algorithms for efficient computation of the generalized $M$-smoother. With the constant-time complexity (per input pixel) filtering algorithms\footnote{In the literature, ``constant time (per pixel)'' is also referred to as ``linear time (in pixel number)''. We follow the way of ``constant time'' in \cite{Chaudhury2011blf}.} \cite{crow1984summed,deriche1992recursively,yang2009real,Chaudhury2011blf,he2010guided}, we are able to compute the weighted averaging in Eq. \eqref{eq.extendedsmoother} in constant time regardless of the size of the neighborhood $\Omega_p$ at each pixel $p$. Note the $\argmin$ step is performed at each pixel individually, thus its computation can be ignored comparing to the filtering step. Let $|\Theta|$ denote the size of set $\Theta$, the computation of Eq. \eqref{eq.ourframework} is mainly the $|\Theta|$ filtering on cost images.


We in this paper mainly target on $8$-bit image, which is the most commonly used format in practice (for $24$-bit or $32$-bit color image, we separately process each of the RGB channels). Thus set $\Theta$ contains $256$ integers in $[0,255]$, \emph{i.e.}, $|\Theta|=256$. We will show how to further reduce the number of filtering in next section.

\begin{algorithm}[t!]
\small
\caption{Approximate Algorithm}
\begin{algorithmic}
\STATE \textbf{Input}: image $I$, number of samples $n$, filter $\mathcal{F}$. \\
\STATE \textbf{Output}: smoothed image $J$.\\
\STATE ---------------------\textbf{Algorithm Start}---------------------
\STATE calculate evenly distributed samples $\hat{\Theta}$ 
\FOR{each sample $\theta$ in $\hat{\Theta}$}
    \STATE (1) compute cost image $D(\theta)=\rho(\theta-I)$;
    \STATE (2) filtering the cost image to get $\mathcal{F}(D(\theta))$;
\ENDFOR
\FOR{each pixel $\mathbf{p}$}
    \STATE (1) compute $\hat{J}_\mathbf{p}$ using Eq. \eqref{eq.approxfirstmin};
    \STATE (2) compute output $J_\mathbf{p}$ using Eq. \eqref{eq.approxfinaloutput};
\ENDFOR
\STATE ----------------------\textbf{Algorithm End}----------------------
\end{algorithmic}
\label{alg:approximatealgo}
\end{algorithm}

\begin{figure*}[!t]
    \vspace{5mm}
    \centerline{\addtocounter{subfigure}{-1}\subfloat{\label{fig.psnrplot.Nl1sig1}
        \includegraphics[width=\swidththree]{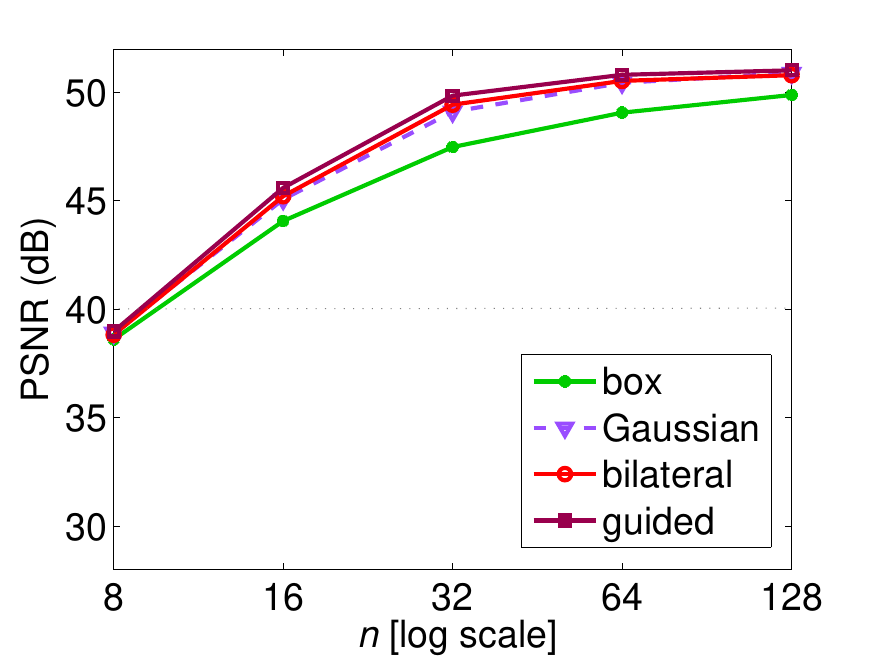}}\hspace{-15pt}
        \hfil \addtocounter{subfigure}{-1}\subfloat{\label{fig.psnrplot.Nl1sig2}
        \includegraphics[width=\swidththree]{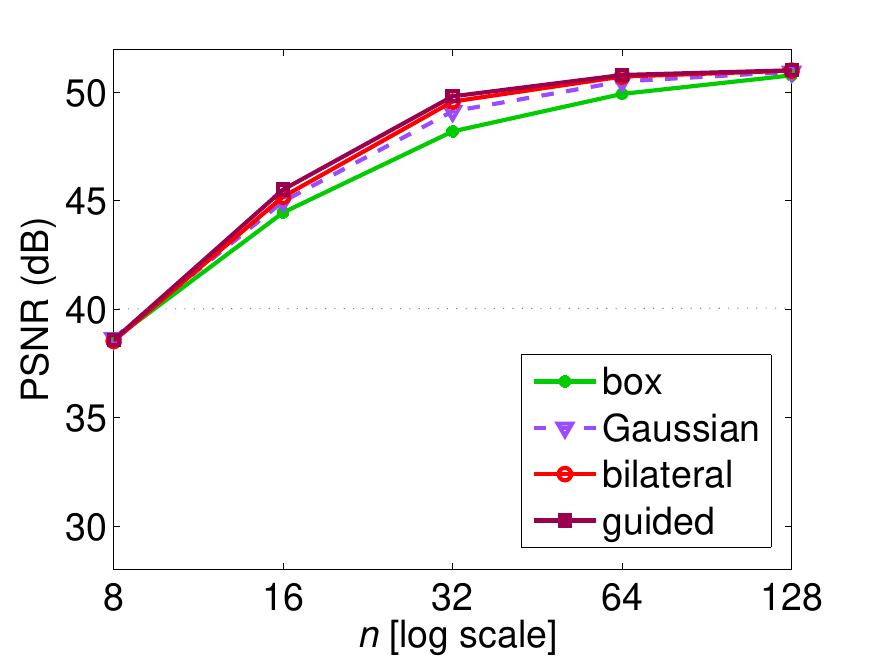}}\hspace{-15pt}
        \hfil \addtocounter{subfigure}{-1}\subfloat{\label{fig.psnrplot.Nl1sig3}
        \includegraphics[width=\swidththree]{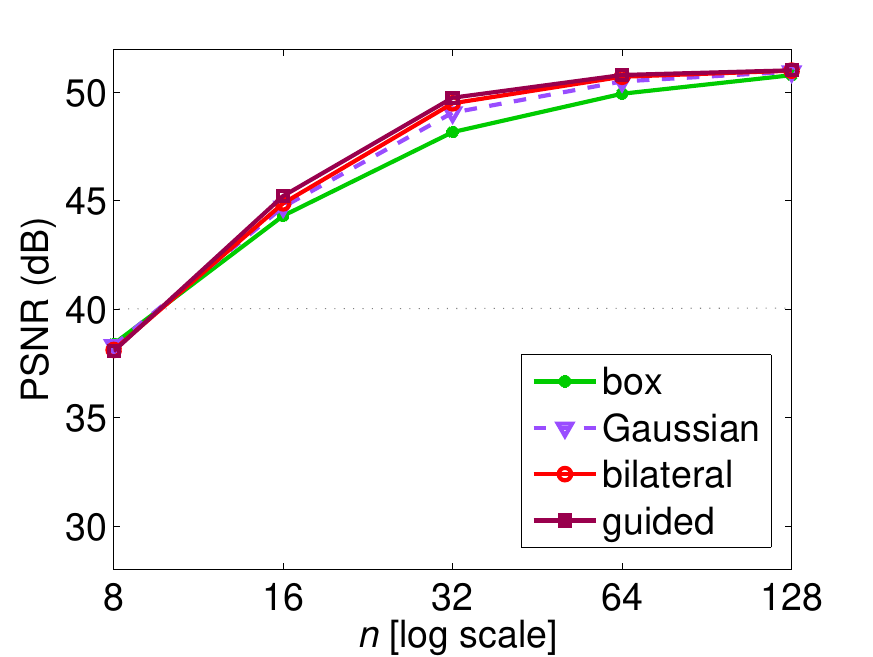}}}\vspace{-10pt}
    \centerline{\addtocounter{subfigure}{-1}\subfloat{\label{fig.psnrplot.Nl1truncsig1}
        \includegraphics[width=\swidththree]{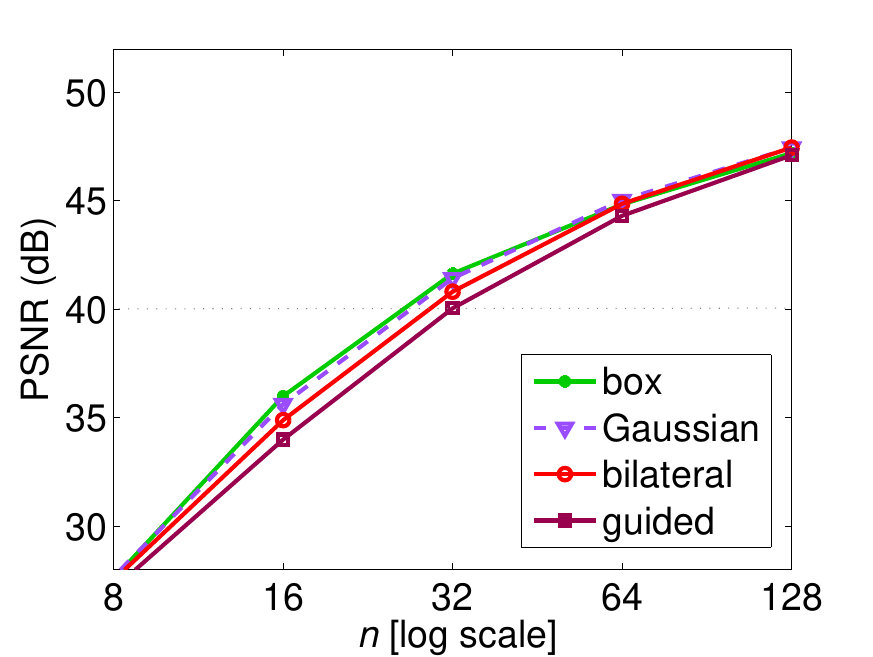}}\hspace{-15pt}
        \hfil \addtocounter{subfigure}{-1}\subfloat{\label{fig.psnrplot.Nl1truncsig2}
        \includegraphics[width=\swidththree]{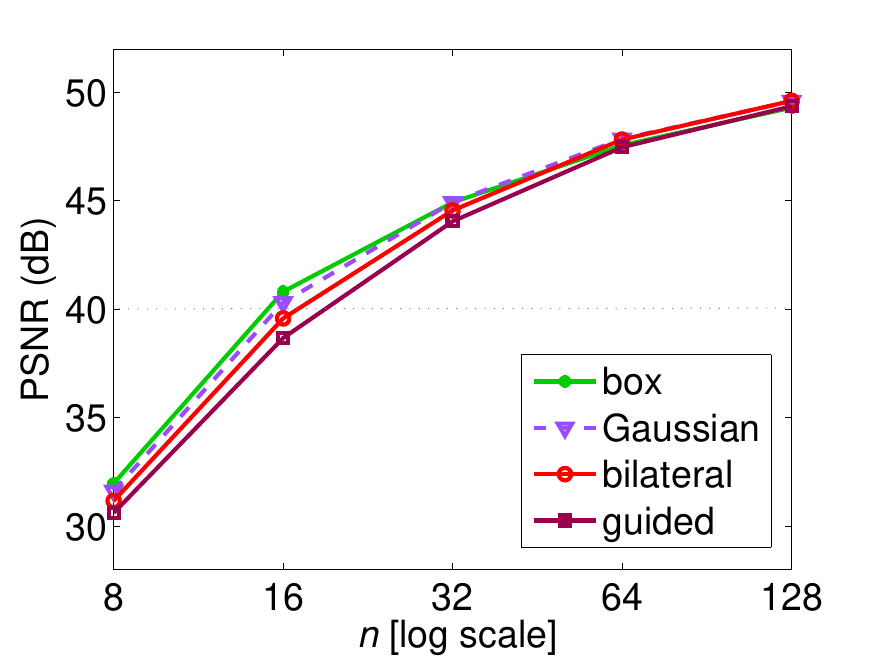}}\hspace{-15pt}
        \hfil \addtocounter{subfigure}{-1}\subfloat{\label{fig.psnrplot.Nl1truncsig3}
        \includegraphics[width=\swidththree]{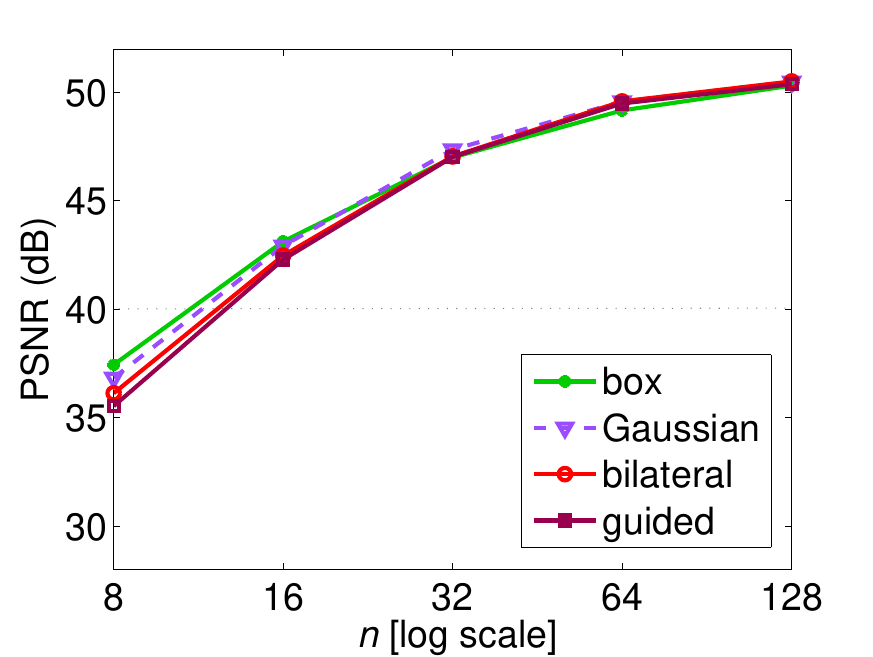}}}\vspace{-10pt}
    \centerline{\addtocounter{subfigure}{-1}\subfloat{\label{fig.psnrplot.Ngausssig1}
        \includegraphics[width=\swidththree]{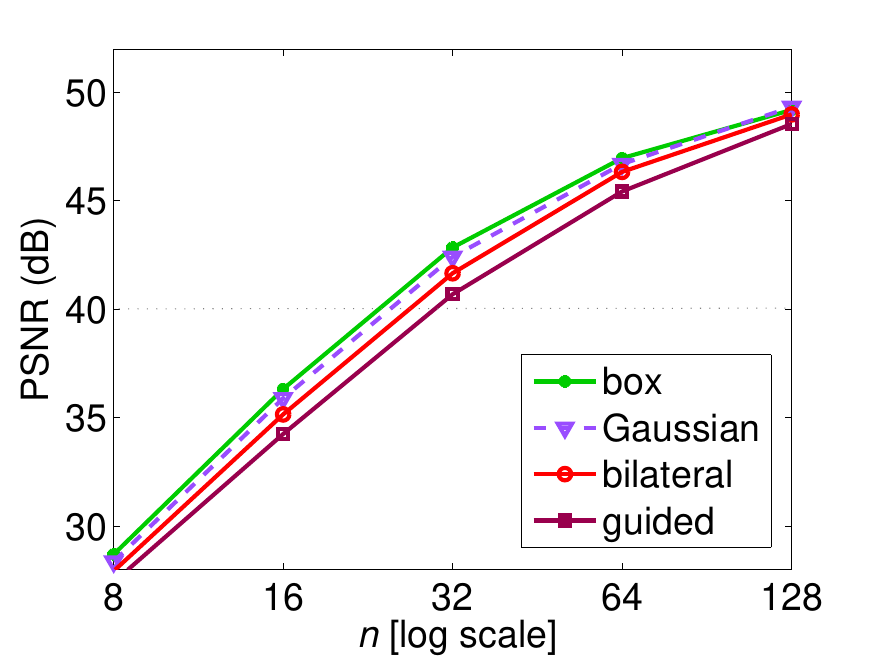}}\hspace{-15pt}
        \hfil \addtocounter{subfigure}{-1}\subfloat{\label{fig.psnrplot.Ngausssig2}
        \includegraphics[width=\swidththree]{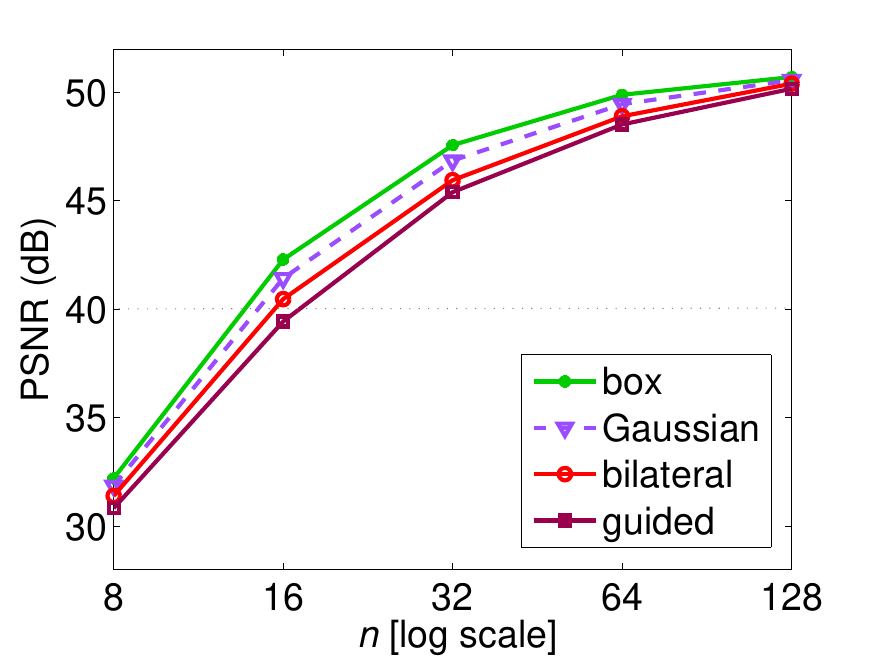}}\hspace{-15pt}
        \hfil \addtocounter{subfigure}{-1}\subfloat{\label{fig.psnrplot.Ngausssig3}
        \includegraphics[width=\swidththree]{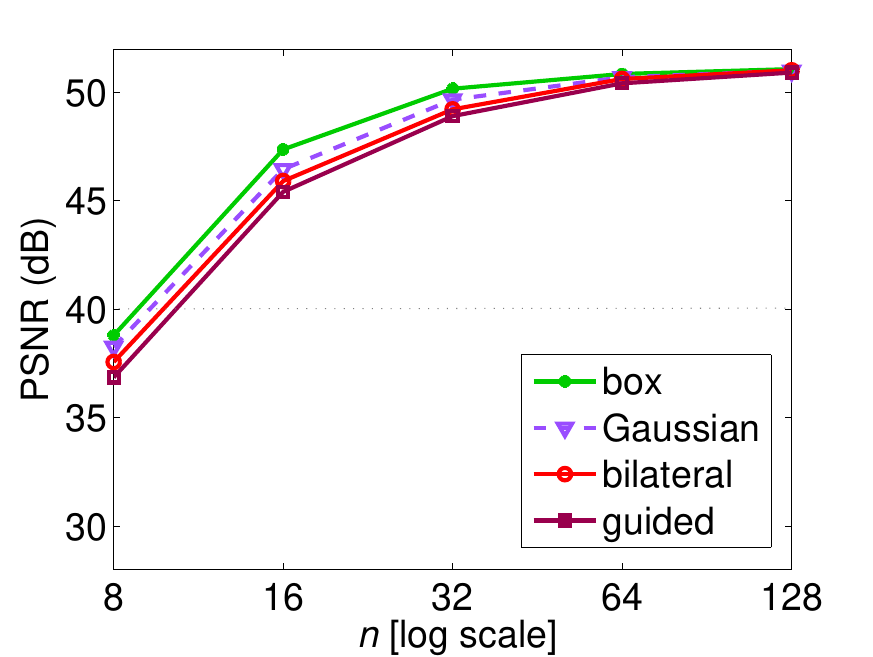}}}\vspace{-10pt}
    \centerline{\addtocounter{subfigure}{-1}\subfloat{\label{fig.psnrplot.Ntukeysig1}
        \includegraphics[width=\swidththree]{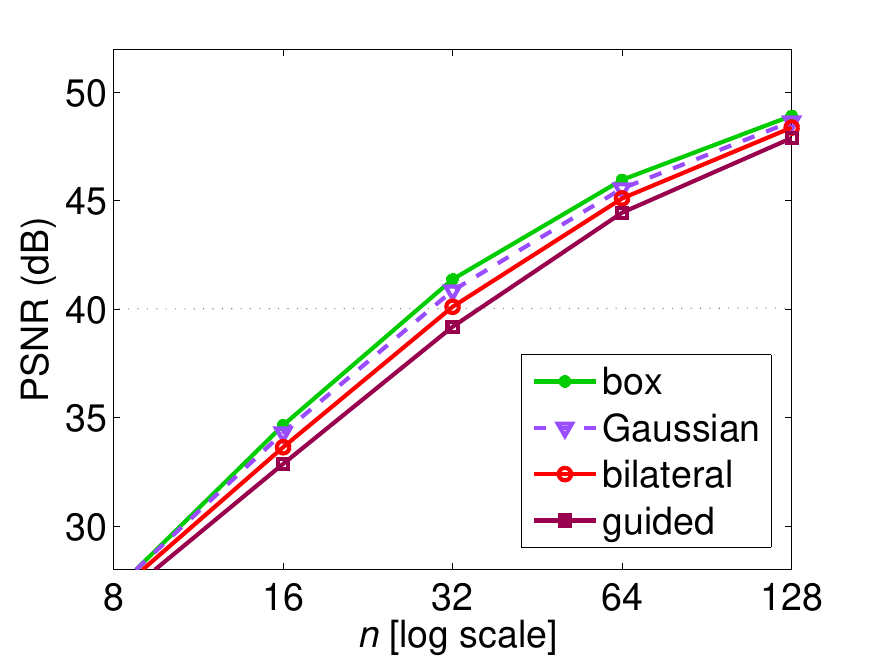}}\hspace{-15pt}
        \hfil \addtocounter{subfigure}{-1}\subfloat{\label{fig.psnrplot.Ntukeysig2}
        \includegraphics[width=\swidththree]{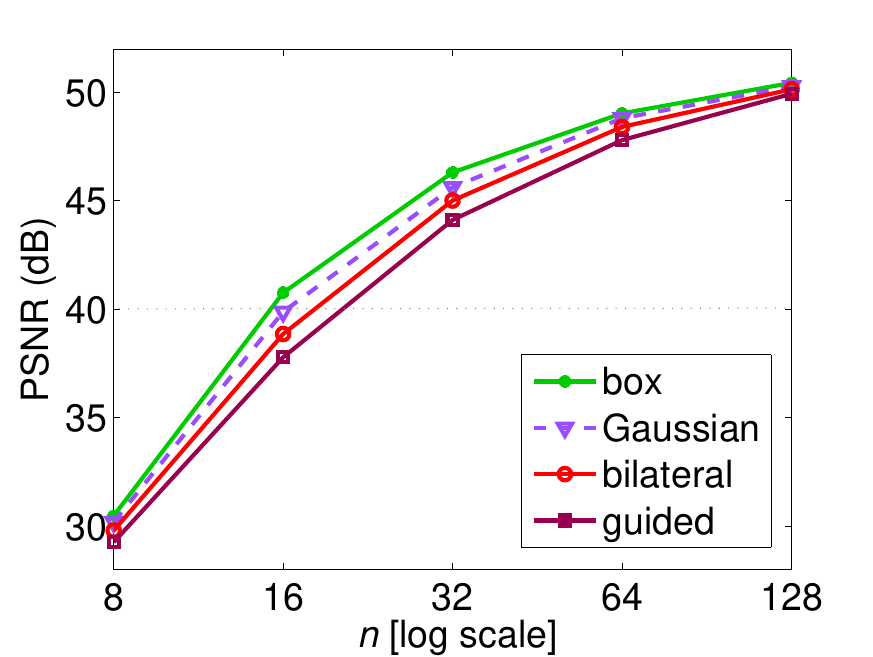}}\hspace{-15pt}
        \hfil \addtocounter{subfigure}{-1}\subfloat{\label{fig.psnrplot.Ntukeysig3}
        \includegraphics[width=\swidththree]{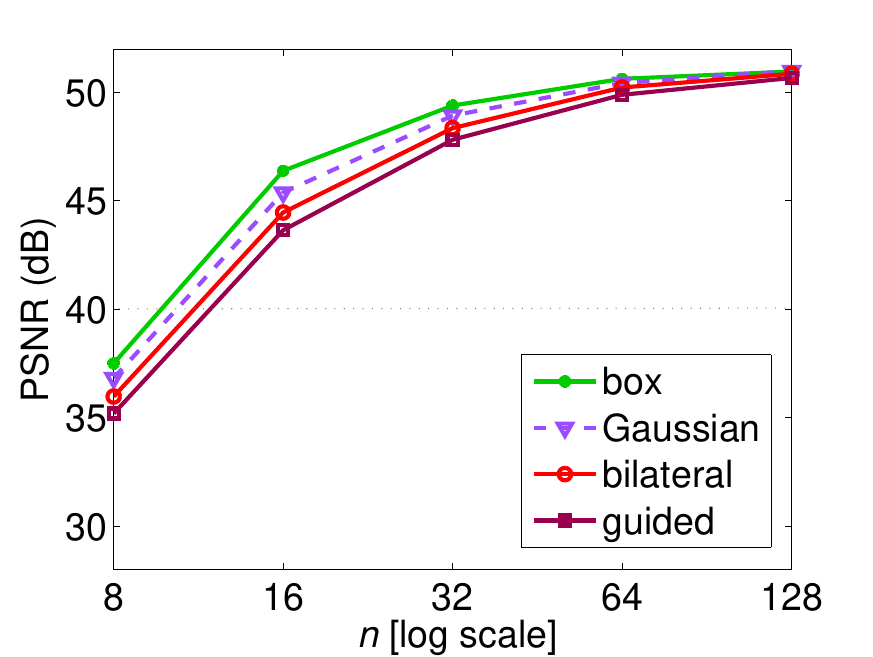}}}\vspace{-10pt}
    \centerline{\subfloat[$\sigma_r=0.05$]{\label{fig.psnrplot.Ngrsig1}
        \includegraphics[width=\swidththree]{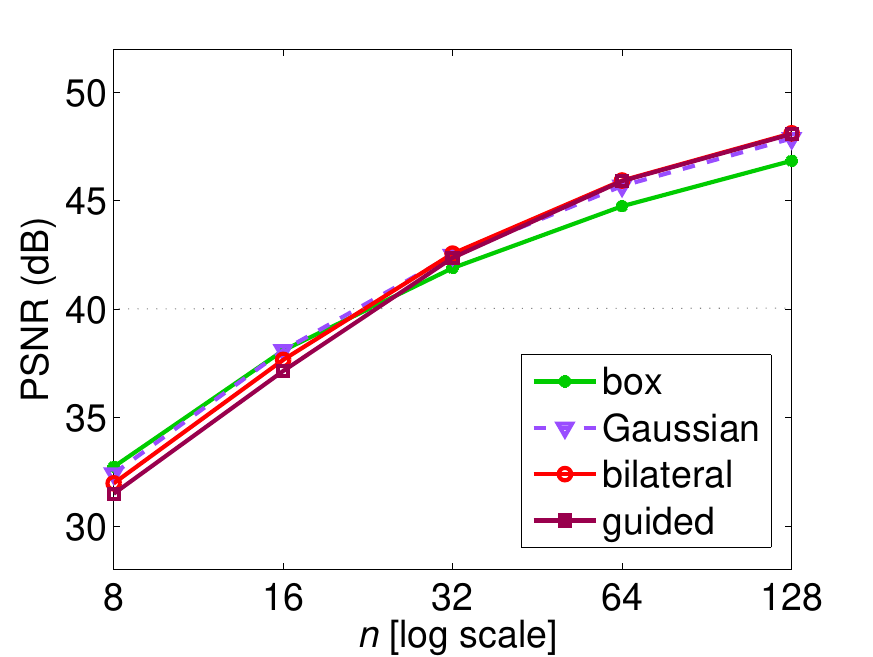}}\hspace{-15pt}
        \hfil \subfloat[$\sigma_r=0.1$]{\label{fig.psnrplot.Ngrsig2}
        \includegraphics[width=\swidththree]{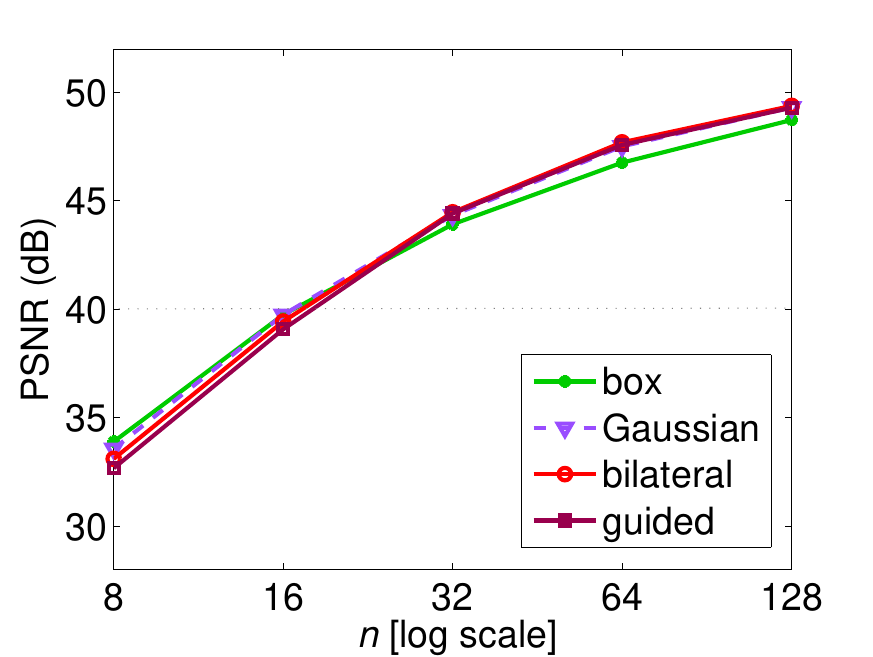}}\hspace{-15pt}
        \hfil \subfloat[$\sigma_r=0.2$]{\label{fig.psnrplot.Ngrsig3}
        \includegraphics[width=\swidththree]{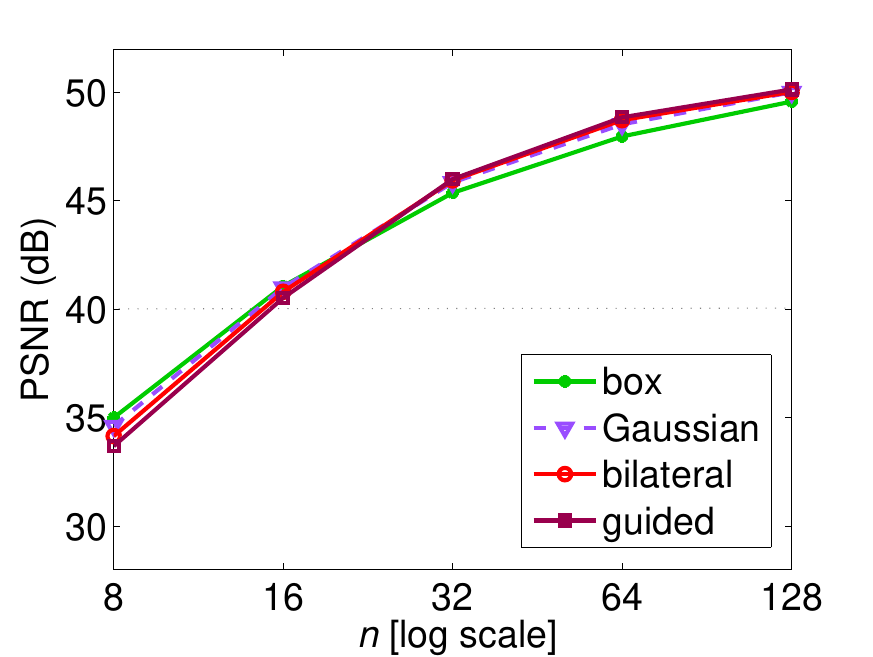}}}
    \caption{PSNR accuracy of the approximate algorithm. The loss functions are (from top to bottom): $L_1$ norm, truncated $L_1$ norm, negative Gauss, Tukey's biweight, and Geman-Reynolds. It is suggested that PSNR value above $40$dB often corresponds to almost invisible differences between two images \protect\cite{paris2009fast}.}
    \label{fig.psnrplot}
\end{figure*}

\subsection{Approximate Algorithm}
An effective strategy to reduce the number of filtering required in the framework is through uniformly sampling the set $\Theta$. The idea is that, filtering is only performed for samples rather than all possible values in $\Theta$, and the output for each pixel is approximated using the samples according to the filtered results (see Algorithm \ref{alg:approximatealgo}). Specifically, let the sampling set be $\hat{\Theta}$, which has $n$ samples evenly distributed in $\Theta$, we first find out the best $\theta$ among these samples at each pixel $\mathbf{p}$
\begin{equation}
    \hat{J}_\mathbf{p} = \argmin_{\theta \in \hat{\Theta}}{\left[\mathcal{F}(D(\theta)) \right]_\mathbf{p}}.
    \label{eq.approxfirstmin}
\end{equation}
Since the following operation is performed at each pixel $\mathbf{p}$ individually, for simplicity, we use $\theta_0$ to denote $\hat{J}_\mathbf{p}$ and $f(\theta)$ to denote the filtered pixel value $\left[\mathcal{F}(D(\theta)) \right]_\mathbf{p}$ for a given $\theta$. For pixel $\mathbf{p}$, suppose $\theta_+$ and $\theta_-$ are the two closest samples near $\theta_0$ in $\hat{\Theta}$, then the output value of pixel $\mathbf{p}$ can be approximated by fitting a parabolic curve \cite{cvpr-07-qingxiong-yang} using the three samples and their corresponding filtered values,
\begin{equation}
    J_\mathbf{p} = \theta_0 - \frac{(\theta_+ - \theta_-)( f(\theta_+) - f(\theta_-))}{4(f(\theta_+) + f(\theta_-) - 2 f(\theta_0))}.
    \label{eq.approxfinaloutput}
\end{equation}
Note that the parabolic fitting is a closed-form approximation by assuming the cost function follows a parabolic curve near the bottom (near $\hat{J}_\mathbf{p}$). The actual cost function cannot be analytically solved without considering the pixel distribution of each pixel's neighborhood due to the data-dependent average (see Appendix for the derivation). Although a more precise way to solve it is to perform further sampling and filtering near $\hat{J}_\mathbf{p}$, we find that the closed-form approximation is much faster and usually accurate enough in practice. We will show this in next section.



\subsection{Experimental Validation}\label{sec:experiments}
As commonly adopted in developing approximate algorithms for bilateral filter \cite{paris2009fast}, we also use the peak signal-to-noise ratio (\textbf{PSNR}) metric to measure the approximate accuracy. Higher PSNR value between the approximate and exact results means more accurate approximation (it is suggested that PSNR value above $40$dB often corresponds to almost invisible differences between two images \cite{paris2009fast}). We test the approximate algorithm for four filters listed in Table \ref{tab:wafall} and five loss functions listed in Table \ref{tab.lossfuncshow} (in total 20 pairs of combination) in the following sampled parameter settings, respectively: $\sigma_s \in \{ 2, 4, 8, 16 \}$, $\sigma_r \in \{ 0.05, 0.1, 0.2, 0.4\}$, and $n \in \{ 8, 16, 32, 64, 128 \}$. The 8 test images are from Paris's bilateral filtering dataset \cite{parisblfdataurl} (color images are converted to grayscale): \emph{dome}, \emph{dragon}, \emph{greekdome}, \emph{housecorner}, \emph{polin}, \emph{swamp}, \emph{tulip}, and \emph{turtle}. The reference image for calculating PSNR is obtained from Eq. \eqref{eq.ourframework} by enumerating all possible output values within $\Theta$ (\emph{i.e.}, integer values in $[0, 255]$ for $8$-bit image).

According to our observation, we find that the PSNR value is not sensitive to the spatial parameter $\sigma_s$. Thus we only plot the PSNR results according to different $\sigma_r$ for each loss function and filter in Fig. \ref{fig.psnrplot}. That is, for a specific pair of loss function and filter, PSNR from different images with different $\sigma_s$ but a same $\sigma_r$ are averaged together for the plot (results for $\sigma_r=0.4$ are not shown since the PSNR are commonly high). The results show that the approximate accuracy for each loss function is sensitive to $\sigma_r$ (except the $L_1$ norm which does not have a scale parameter). Commonly speaking, the larger value of $\sigma_r$, the smaller $n$ required for high accuracy (PSNR above $40$dB). For example, for $\sigma_r=0.05$ (first row), $n=32$ can commonly make the PSNR above $40$dB, while for $\sigma_r=0.1$ (second row), $n=16$ is often enough.

The above experimental results imply that in practice we can safely use a $n$ much smaller than 256 (for 8-bit image). Generally, for situations where parameter $\sigma_r$ is not too small (e.g., $\geq 0.1$), we can use $n=16$ to get accurately approximated results ($n$ can be further reduced for larger $\sigma_r$).

\begin{table}[!t]
 \footnotesize
 \center
 \caption{Approximating histogram filters \label{tab:acchistfilter}}
 \begin{tabular}{r|cc}
  \toprule
  & $\mathcal{F}$ is box filter & $\mathcal{F}$ is Gaussian filter \\
  \hline
  $L_1$ norm & median filter  & isotropic median filter \cite{kass2010smoothed}  \\
  RILF$^\dag$ & global mode filter$^\ddag$ \cite{van2001local} & dominant-mode filter$^\ddag$ \cite{kass2010smoothed} \\
  \bottomrule
  \multicolumn{3}{l}{\hspace{-6pt}\scriptsize{$\dag$: RILF stands for redescending-influence loss functions.}} \\
  \multicolumn{3}{l}{\multirow{3}{0.90\linewidth}{\hspace{-6pt}\scriptsize{$\ddag$: The global mode filter uses hard spatial window to construct local histogram, while the dominant-mode filter uses Gaussian weighted window to construct local histogram.}}} \\ \multicolumn{3}{l}{ } \\  \multicolumn{3}{l}{ } \\ \multicolumn{3}{l}{ } \\
 \end{tabular}
 \begin{tabular}{c||c|c||c|c}
 \multicolumn{5}{l}{\scriptsize{Timing (per mega-pixel) with $n=16$ (RILF is truncated $L_1$ norm):}}\\
 \toprule
 filter & \multicolumn{2}{c||}{CPU$^\dag$} & \multicolumn{2}{c}{GPU$^\dag$} \\
 type & ours & \cite{kass2010smoothed} & ours & \cite{kass2010smoothed} \\
 \hline
 median            & 150 ms  & --      & 5 ms & -- \\
 isotropic median  & 230 ms  & 833 ms  & 6 ms & 166 ms \\
 global mode       & 150 ms  & --      & 5 ms & -- \\
 dominant-mode     & 230 ms  & 2777 ms & 6 ms & 332 ms \\
 \bottomrule
 \multicolumn{5}{l}{\multirow{6}{0.90\linewidth}{\hspace{-6pt}\scriptsize{$\dag$: The timing of \cite{kass2010smoothed} is reproduced from the paper. It is reported on Intel 2.83 GHz Xeon E5440 CPU and NVIDIA Quadro FX 770M GPU. Note that our algorithm for dominant-mode filter only needs half as many Gaussian filtering as \cite{kass2010smoothed} and is much simpler to implement (the algorithm in \cite{kass2010smoothed} requires $2n$ Gaussian filtering for computing integrals and derivatives of histogram).}}} \\ \multicolumn{5}{l}{ } \\ \multicolumn{5}{l}{ } \\ \multicolumn{5}{l}{ } \\ \multicolumn{5}{l}{ } \\
 \end{tabular}
\end{table}

\begin{figure}[!t]
    \centerline{\subfloat[Input]{\label{fig.domimodecompare.sigr21}
    \includegraphics[width=\swidththree]{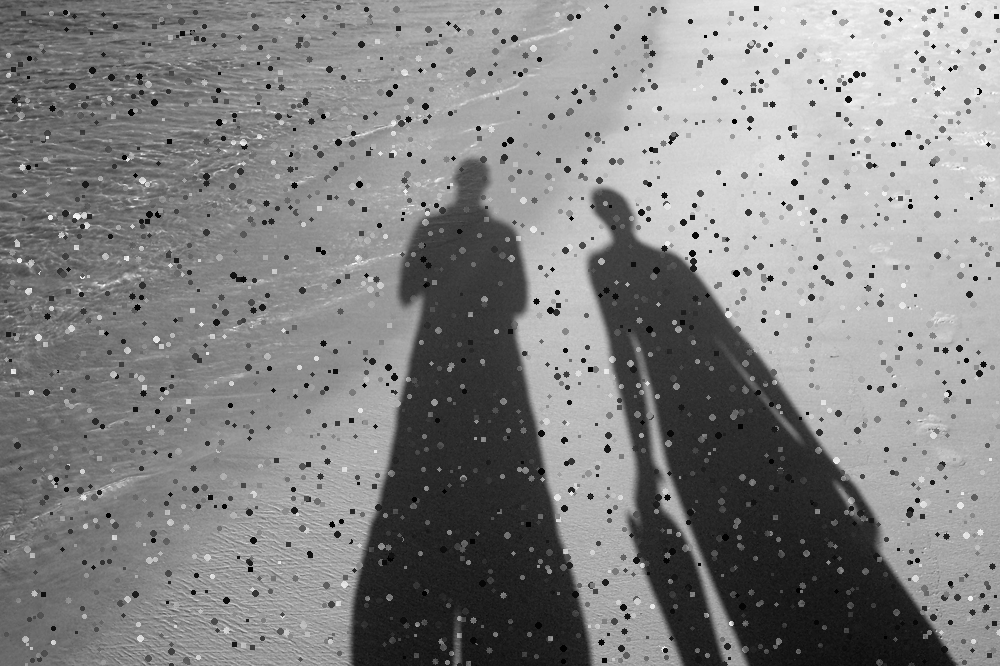}}\hspace{-6pt}
    \hfil \subfloat[Ours ($\mathcal{F}$ is BF)]{\label{fig.domimodecompare.sigr22}
    \includegraphics[width=\swidththree]{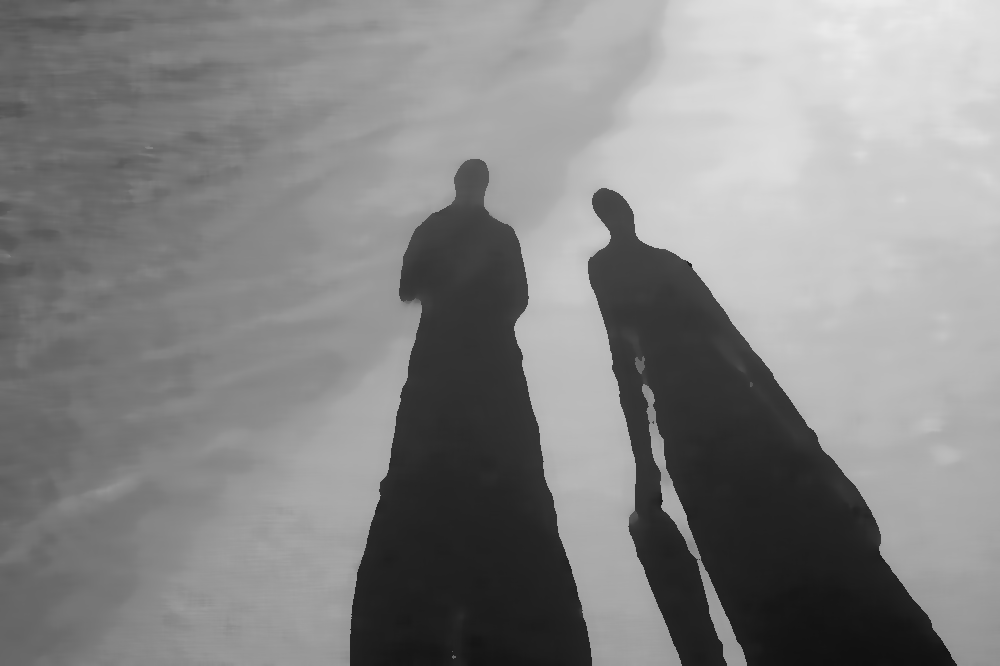}}\hspace{-6pt}
    \hfil \subfloat[Ours ($\mathcal{F}$ is GF)]{\label{fig.domimodecompare.sigr23}
    \includegraphics[width=\swidththree]{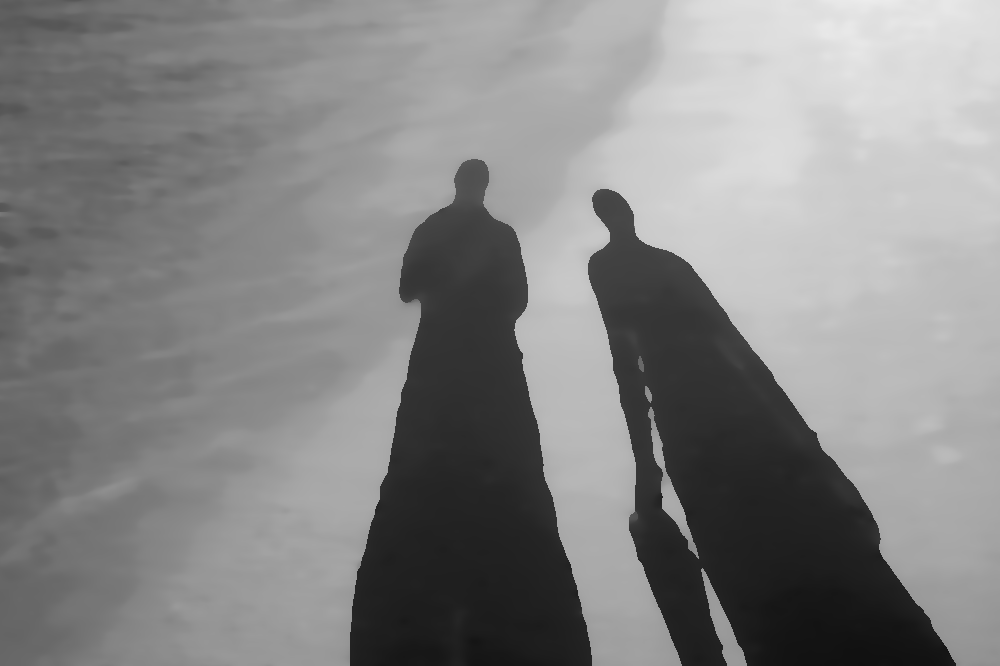}}}
    \caption{Example result of our approximate global mode filter ($\mathcal{F}$ is box filter) and dominant-mode filter ($\mathcal{F}$ is Gaussian filter).}
    \label{fig.domimodecompare}
\end{figure}

\begin{figure*}[!t]
    \vspace{-3mm}
    \centerline{\subfloat[BLF ($\sigma_r=0.1$)]{\label{fig.enhancebfandgf.Nbox}
        \begin{minipage}[b]{\swidthfour}
        \includegraphics[width=\swidthone]{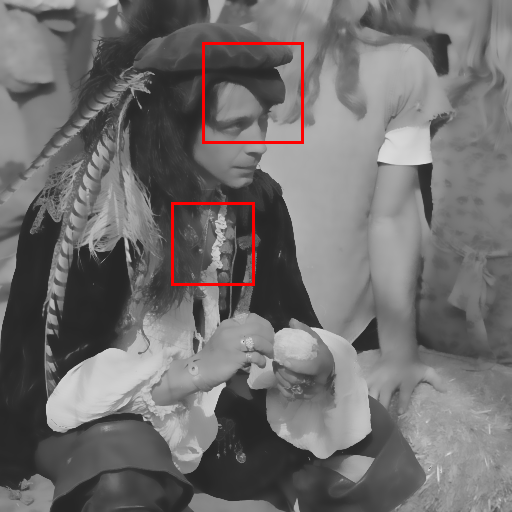}\vspace{-8pt}
        \centerline{\addtocounter{subfigure}{-1}\subfloat{
            \includegraphics[width=\swidthtwo]{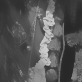}~%
            \includegraphics[width=\swidthtwo]{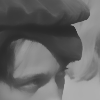}%
        }}
        \end{minipage}}
        \hfil \subfloat[BLF ($\sigma_r=0.2$)]{\label{fig.enhancebfandgf.Ngf}
        \begin{minipage}[b]{\swidthfour}
        \includegraphics[width=\swidthone]{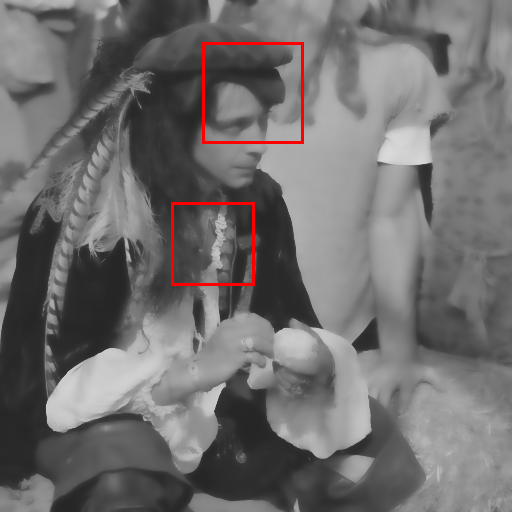}\vspace{-8pt}
        \centerline{\addtocounter{subfigure}{-1}\subfloat{
            \includegraphics[width=\swidthtwo]{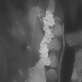}~%
            \includegraphics[width=\swidthtwo]{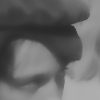}%
        }}
        \end{minipage}}
        \hfil \subfloat[BLF with $L_1$ norm]{\label{fig.enhancebfandgf.Ngauss}
        \begin{minipage}[b]{\swidthfour}
        \includegraphics[width=\swidthone]{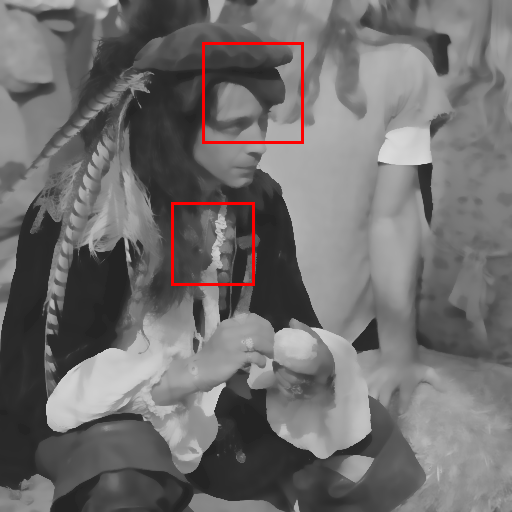}\vspace{-8pt}
        \centerline{\addtocounter{subfigure}{-1}\subfloat{
            \includegraphics[width=\swidthtwo]{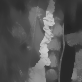}~%
            \includegraphics[width=\swidthtwo]{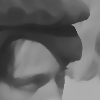}%
        }}
        \end{minipage}}
        \hfil \subfloat[BLF with truncated $L_1$]{\label{fig.enhancebfandgf.gemanreynolds}
        \begin{minipage}[b]{\swidthfour}
        \includegraphics[width=\swidthone]{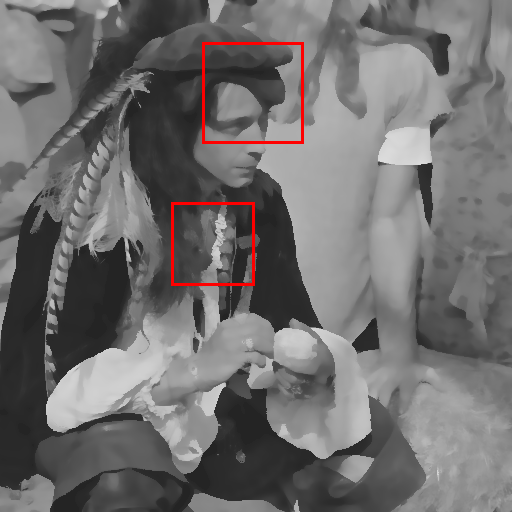}\vspace{-8pt}
        \centerline{\addtocounter{subfigure}{-1}\subfloat{
            \includegraphics[width=\swidthtwo]{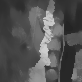}~%
            \includegraphics[width=\swidthtwo]{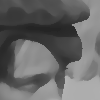}%
        }}
        \end{minipage}}}\vspace{-8pt}
    \centerline{\subfloat[GDF ($\sigma_r=0.1$)]{\label{fig.enhancebfandgf.alphabox}
        \begin{minipage}[b]{\swidthfour}
        \includegraphics[width=\swidthone]{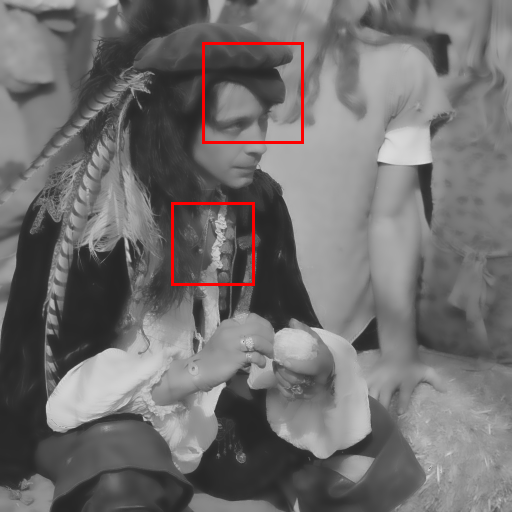}\vspace{-8pt}
        \centerline{\addtocounter{subfigure}{-1}\subfloat{
            \includegraphics[width=\swidthtwo]{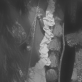}~%
            \includegraphics[width=\swidthtwo]{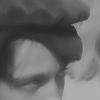}%
        }}
        \end{minipage}}
        \hfil \subfloat[GDF ($\sigma_r=0.2$)]{\label{fig.enhancebfandgf.alphagf}
        \begin{minipage}[b]{\swidthfour}
        \includegraphics[width=\swidthone]{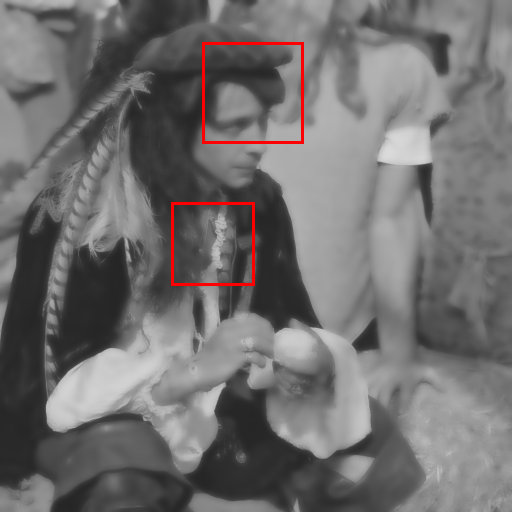}\vspace{-8pt}
        \centerline{\addtocounter{subfigure}{-1}\subfloat{
            \includegraphics[width=\swidthtwo]{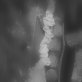}~%
            \includegraphics[width=\swidthtwo]{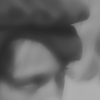}%
        }}
        \end{minipage}}
        \hfil \subfloat[GDF with $L_1$ norm]{\label{fig.enhancebfandgf.alphagauss}
        \begin{minipage}[b]{\swidthfour}
        \includegraphics[width=\swidthone]{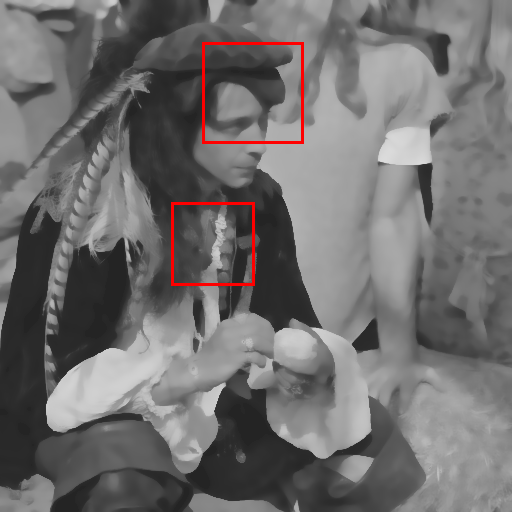}\vspace{-8pt}
        \centerline{\addtocounter{subfigure}{-1}\subfloat{
            \includegraphics[width=\swidthtwo]{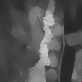}~%
            \includegraphics[width=\swidthtwo]{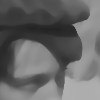}%
        }}
        \end{minipage}}
        \hfil \subfloat[GDF with truncated $L_1$]{\label{fig.enhancebfandgf.alphablf}
        \begin{minipage}[b]{\swidthfour}
        \includegraphics[width=\swidthone]{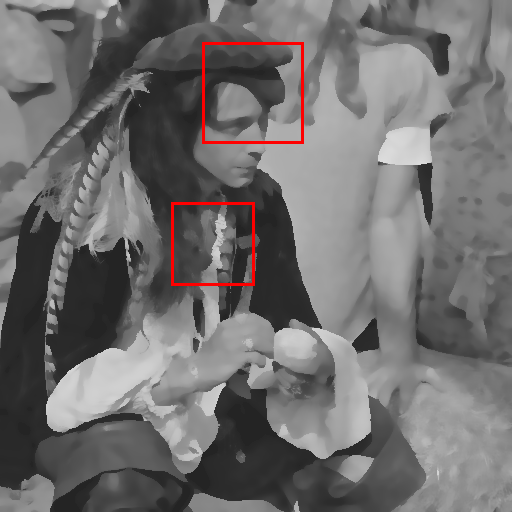}\vspace{-8pt}
        \centerline{\addtocounter{subfigure}{-1}\subfloat{
            \includegraphics[width=\swidthtwo]{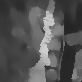}~%
            \includegraphics[width=\swidthtwo]{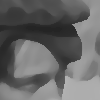}%
        }}
        \end{minipage}}}
    \caption{Piecewise-constant smoothing with proposed framework. Spatial parameter is $\sigma_s=3$ for all results. The $\sigma_r$ in our framework is $0.2$. The bilateral filter and guided filter with $\sigma_r=0.1$ cannot smooth out fine details (see left close-up window), while with larger parameter $\sigma_r=0.2$ they may blur major edges (see right close-up window).}
    \label{fig.enhancebfandgf}
    \vspace{-3mm}
\end{figure*}

\section{Exploiting Specific Forms}\label{sec:analysis}

We in this section demonstrate our contribution by exploiting the specific forms of the generalized $M$-smoother and our filtering framework. Since during our previous experiments, we find that the smoothing effects of the four redescending-influence loss functions are actually visually similar to each other (but different from the $L_1$ norm). We will use one of the last four loss functions as a representative when demonstrating the filtering effects in this section.

\begin{figure}[!t]
    \centerline{\subfloat[Input]{\label{fig.compmodegf.sigr21}
    \includegraphics[width=\swidththree]{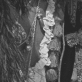}}\hspace{-4pt}
    \hfil \subfloat[Ours ($\mathcal{F}$ is GF)]{\label{fig.compmodegf.sigr22}
    \includegraphics[width=\swidththree]{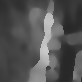}}\hspace{-4pt}
    \hfil \subfloat[Ours ($\mathcal{F}$ is GDF)]{\label{fig.compmodegf.sigr23}
    \includegraphics[width=\swidththree]{images/pirate.png_sigs3_loss4_sigr0.20_ft3_nr32_psnr0.00_small2}}}
    \caption{Comparison between Gaussian filter and guided filter as $\mathcal{F}$ (with truncated $L_1$ norm, $\sigma_s=3$, $\sigma_r=0.2$). The comparison shows that, when $\mathcal{F}$ is an edge-preserving filter (GDF), the framework can better preserve edges, while on the other hand, when $\mathcal{F}$ is a linear filter (GF), it can achieve stronger smoothing but may cause deviations from input edges.}
    \label{fig.compmodegf}
\end{figure}

\begin{figure}[!t]
    \centerline{\subfloat[Input]{\label{fig.noisesyn.ori}
        \includegraphics[width=\swidththree]{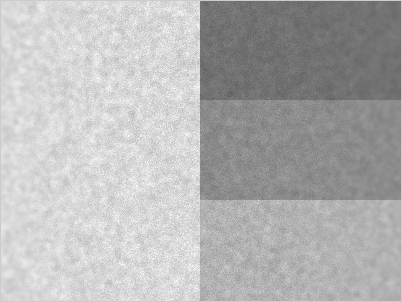}}\hspace{-6pt}
        \hfil \subfloat[BLF]{\label{fig.noisesyn.blf}
        \includegraphics[width=\swidththree]{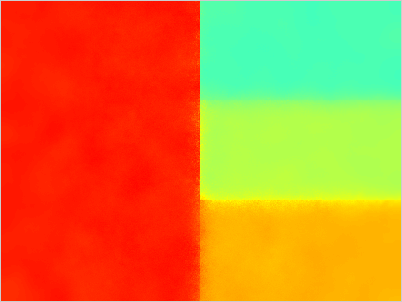}}\hspace{-6pt}
        \hfil \subfloat[GDF]{\label{fig.noisesyn.gf}
        \includegraphics[width=\swidththree]{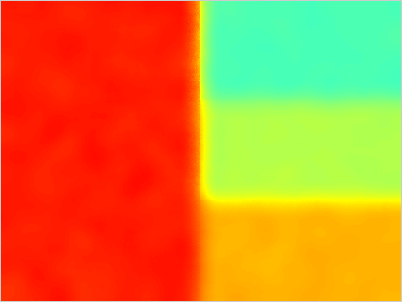}}}\vspace{-8pt}
    \vfil \centerline{\subfloat[Input (visualized)]{\label{fig.noisesyn.oricolor}
        \includegraphics[width=\swidththree]{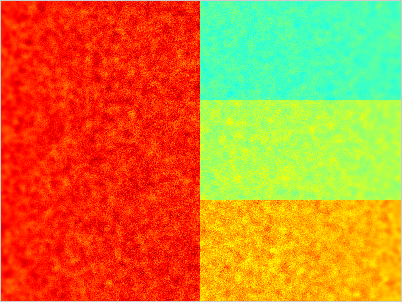}}\hspace{-6pt}
        \hfil \subfloat[Ours ({$\mathcal{F}$} is BLF)]{\label{fig.noisesyn.ourblf}
        \includegraphics[width=\swidththree]{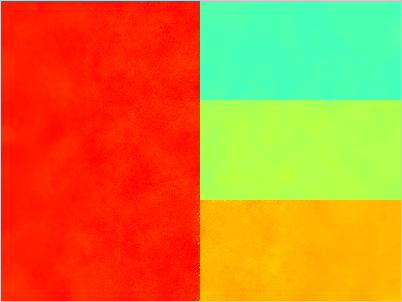}}\hspace{-6pt}
        \hfil \subfloat[Ours ({$\mathcal{F}$} is GDF)]{\label{fig.noisesyn.ourgf}
        \includegraphics[width=\swidththree]{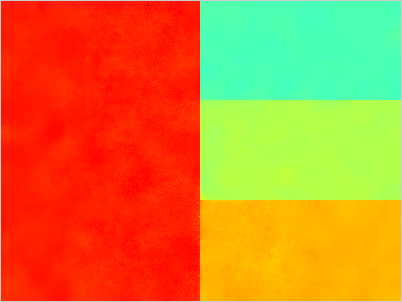}}}
    \caption{Smoothing of a synthetic grayscale noisy image. The colored visualized display is shown for clarity. The parameters used in BLF and GDF are $\sigma_s=10$, $\sigma_r=0.15$. The loss function in our framework is truncated $L_1$ norm.}
    \label{fig.noisesyn}
\end{figure}

\begin{figure*}[!t]
    \centerline{\subfloat[Clean RGB image]{\label{fig.depthdenoising.rgb}
            \begin{minipage}[b]{0.19\linewidth}
            \centerline{\includegraphics[width=\swidthone]{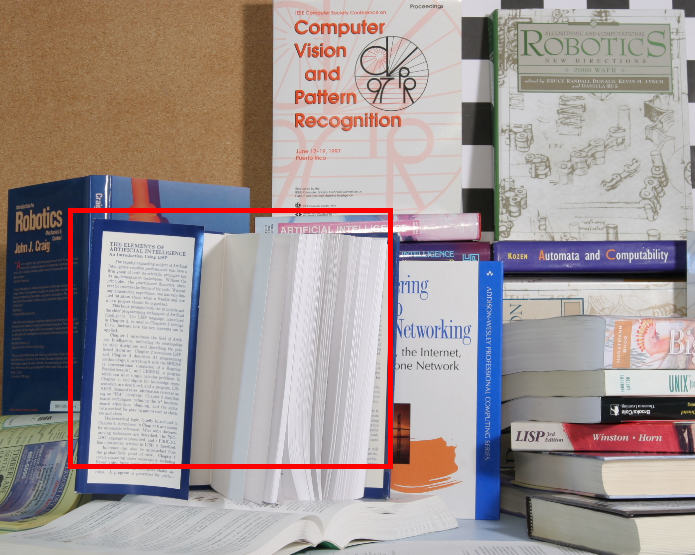}}
            \centerline{\includegraphics[width=\swidthone]{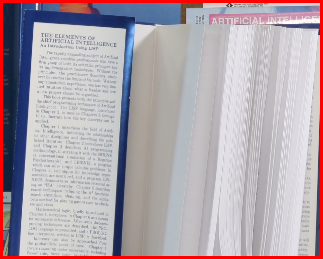}}
            \end{minipage}
        }\hspace{-6pt}
        \hfil \subfloat[Ground truth]{\label{fig.depthdenoising.gt}
            \begin{minipage}[b]{0.19\linewidth}
            \centerline{\includegraphics[width=\swidthone]{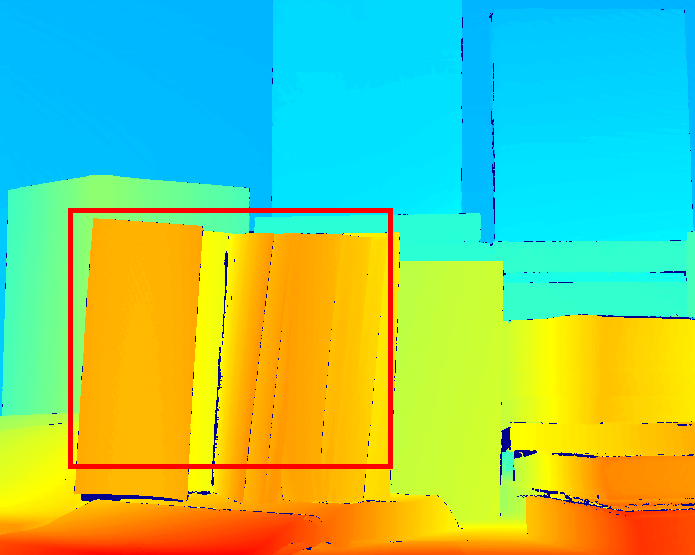}}
            \centerline{\includegraphics[width=\swidthone]{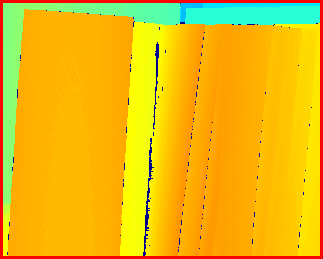}}
            \end{minipage}
        }\hspace{-6pt}
        \hfil \subfloat[With noise($74.6\%$)]{\label{fig.depthdenoising.noisy}
            \begin{minipage}[b]{0.19\linewidth}
            \centerline{\includegraphics[width=\swidthone]{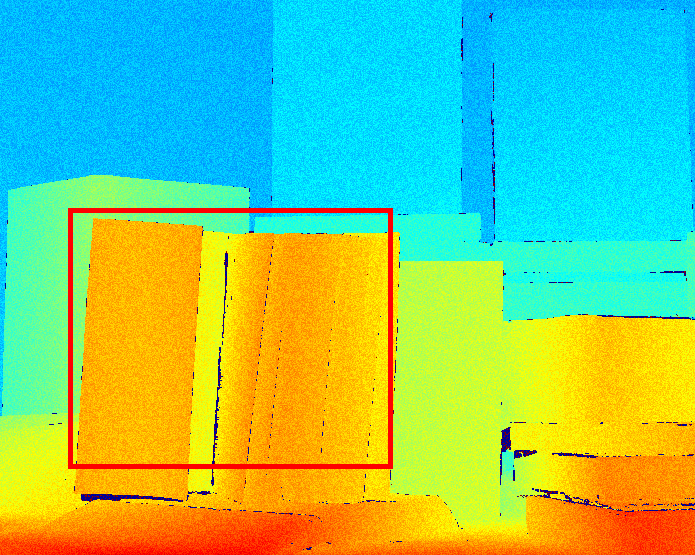}}
            \centerline{\includegraphics[width=\swidthone]{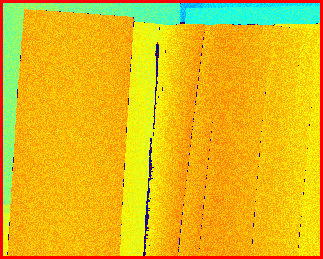}}
            \end{minipage}
        }\hspace{-6pt}
        \hfil \subfloat[Joint BLF ($13.0\%$)]{\label{fig.depthdenoising.jblf}
            \begin{minipage}[b]{0.19\linewidth}
            \centerline{\includegraphics[width=\swidthone]{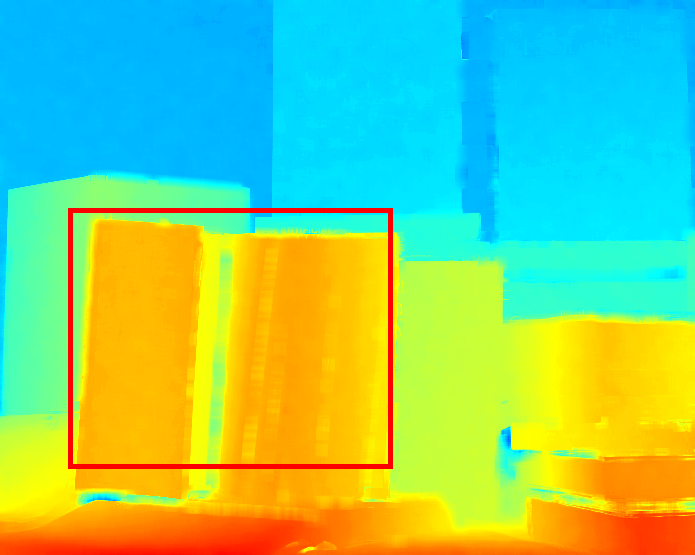}}
            \centerline{\includegraphics[width=\swidthone]{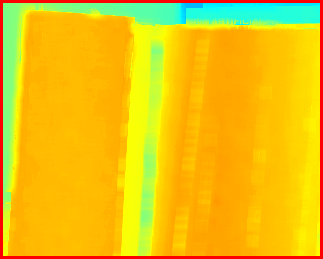}}
            \end{minipage}
        }\hspace{-6pt}
        \hfil \subfloat[Ours ($4.29\%$)]{\label{fig.depthdenoising.our}
            \begin{minipage}[b]{0.19\linewidth}
            \centerline{\includegraphics[width=\swidthone]{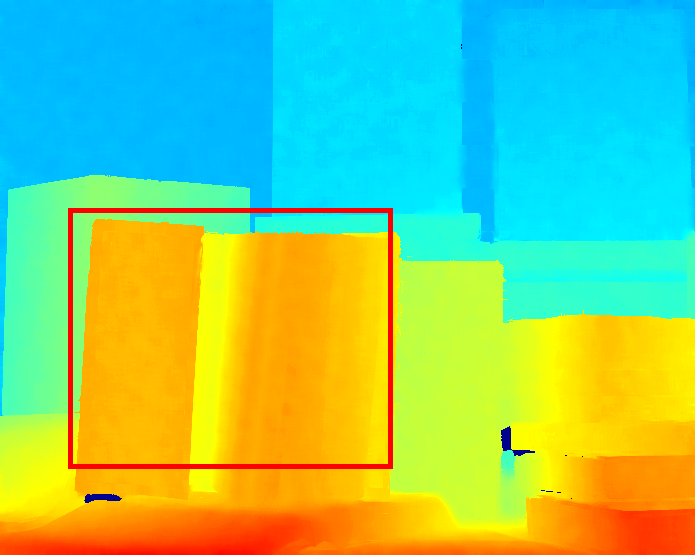}}
            \centerline{\includegraphics[width=\swidthone]{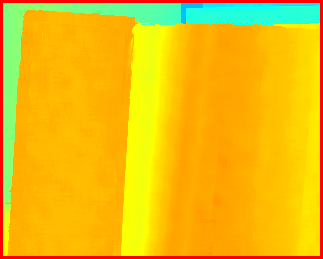}}
            \end{minipage}
        }}\vspace{-8pt}
    \centerline{\subfloat[Clean RGB image]{\label{fig.depthdenoising.rgb1}
            \begin{minipage}[b]{0.19\linewidth}
            \centerline{\includegraphics[width=\swidthone]{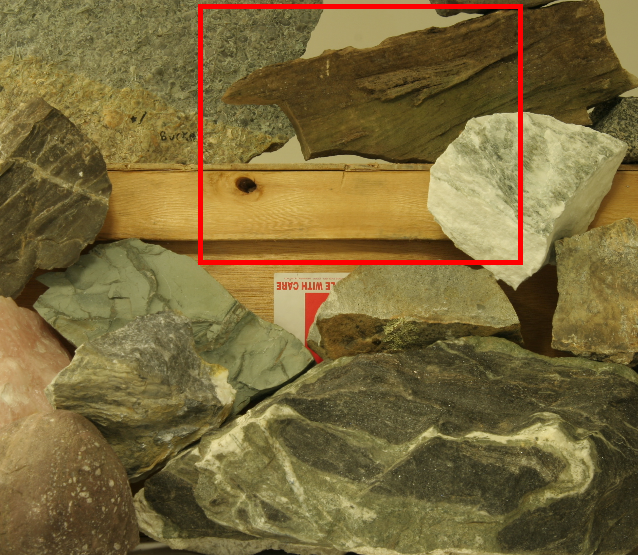}}
            \centerline{\includegraphics[width=\swidthone]{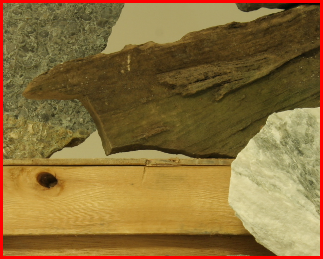}}
            \end{minipage}
        }\hspace{-6pt}
        \hfil \subfloat[Ground truth]{\label{fig.depthdenoising.gt1}
            \begin{minipage}[b]{0.19\linewidth}
            \centerline{\includegraphics[width=\swidthone]{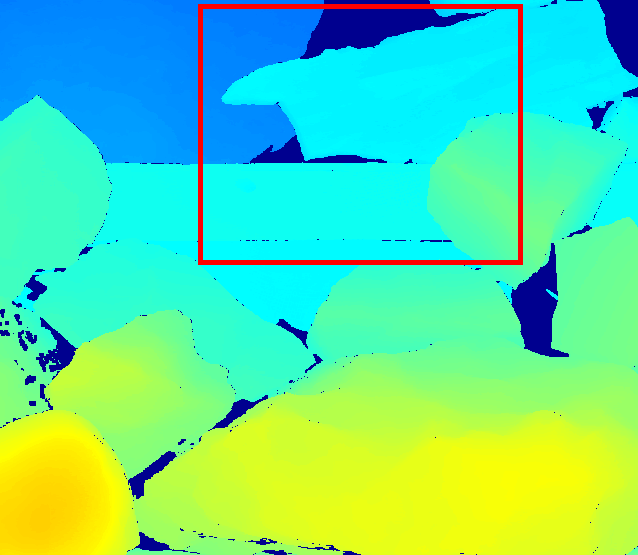}}
            \centerline{\includegraphics[width=\swidthone]{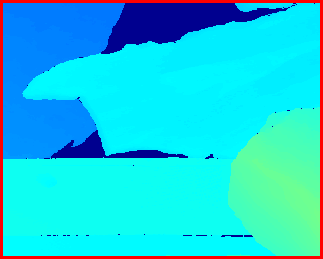}}
            \end{minipage}
        }\hspace{-6pt}
        \hfil \subfloat[With noise($73.2\%$)]{\label{fig.depthdenoising.noisy1}
            \begin{minipage}[b]{0.19\linewidth}
            \centerline{\includegraphics[width=\swidthone]{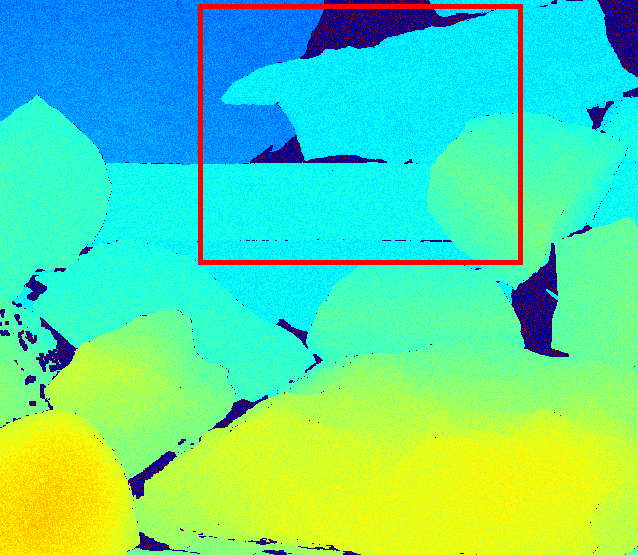}}
            \centerline{\includegraphics[width=\swidthone]{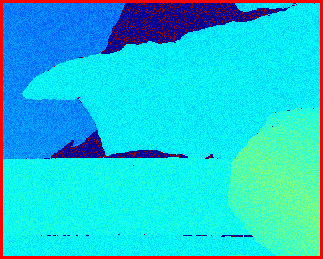}}
            \end{minipage}
        }\hspace{-6pt}
        \hfil \subfloat[Joint BLF ($13.3\%$)]{\label{fig.depthdenoising.jblf1}
            \begin{minipage}[b]{0.19\linewidth}
            \centerline{\includegraphics[width=\swidthone]{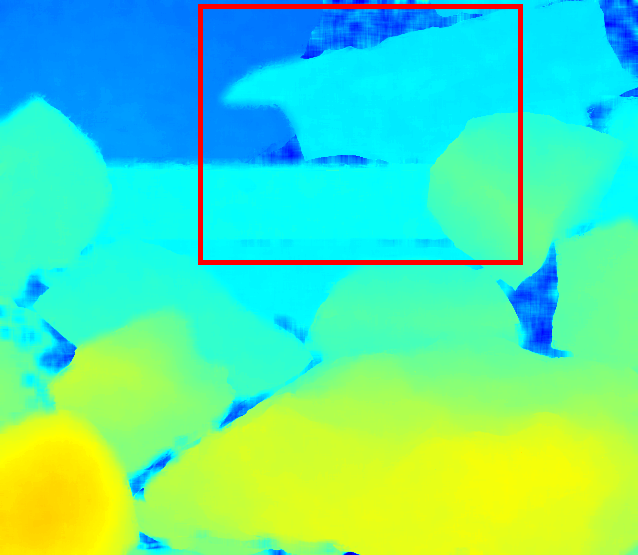}}
            \centerline{\includegraphics[width=\swidthone]{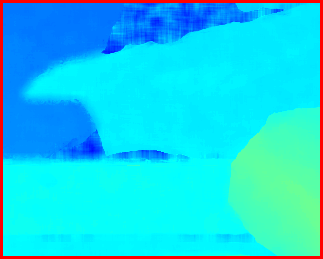}}
            \end{minipage}
        }\hspace{-6pt}
        \hfil \subfloat[Ours ($3.45\%$)]{\label{fig.depthdenoising.our1}
            \begin{minipage}[b]{0.19\linewidth}
            \centerline{\includegraphics[width=\swidthone]{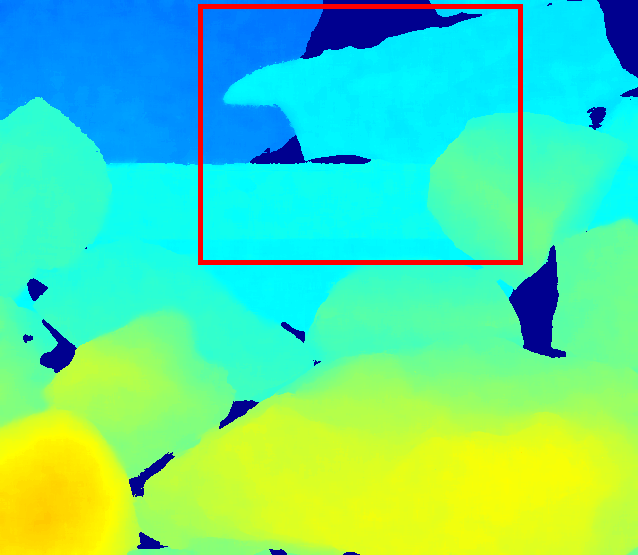}}
            \centerline{\includegraphics[width=\swidthone]{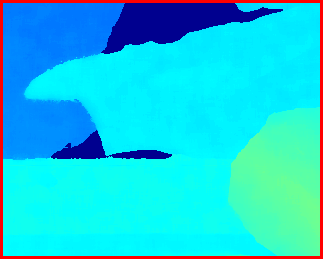}}
            \end{minipage}
        }}
    \caption{Joint filtering for disparity map denoising. From left to right: clean RGB images, ground-truth disparity maps, disparity maps deteriorated with Gaussian noise, denoised disparity maps using joint bilateral filter, denoised disparity maps using our enhanced joint bilateral filter with truncated $L_1$ norm as loss function. The parameters are $\sigma_s=5$, $\sigma_r=0.1$. The percentage shown under subfigures is the the percentage of bad estimated pixels: we adopted the methodology used in \cite{scharstein2002taxonomy}: if the disparity error of a pixel is larger than $1$, it is treated as a bad pixel. Notice that denoising on natural/textured images \cite{zuo2014gradienttip} or based on complicated noise model \cite{jiang14mixednoise} is out of the scope of this paper.}
    \label{fig.depthdenoising}
\end{figure*}

\subsection{Fast Algorithms for Histogram Filters}\label{sec:specificlinear}
As discussed in Sec. \ref{sec:relatedworkrobust}, traditional $M$-smoother is closely related to histogram filters. Thus the proposed framework with $\mathcal{F}$ being box filter or Gaussian filter can serve as fast approximation for histogram filters. Table \ref{tab:acchistfilter} shows the correspondence between proposed framework and histogram filters.

Note that both box filter and Gaussian filter can be implemented in constant time complexity (per input pixel) \cite{crow1984summed,deriche1992recursively}. In our implementation, box filter takes 5 milliseconds per mega-pixel (\textbf{ms/Mp}) on CPU and 0.25 ms/Mp on GPU, while the Gaussian filter takes 12 ms/Mp on CPU and 0.3 ms/Mp on GPU\footnote{The CPU running time is obtained on Intel 3.4 GHz Core i7-3770 CPU with 8GB RAM, using single thread implementation. The GPU running time is obtained on a NVIDIA GTX 780 graphics card, using CUDA implementation.}. The running time of our approximated local-histogram-based filters is shown in Table \ref{tab:acchistfilter}. Fig. \ref{fig.domimodecompare} shows an example result of our approximate global mode filter and dominant-mode filter.



\begin{figure*}[!t]
    \centerline{\subfloat[Input RGB]{\label{fig.kinectdepth.ori}
        \includegraphics[width=\swidthfour]{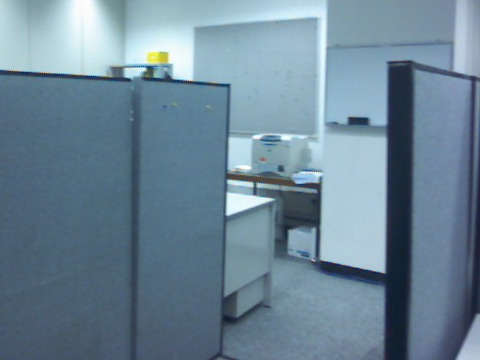}}\hspace{-6pt}
        \hfil \subfloat[Input depth]{\label{fig.kinectdepth.oridepth}
        \includegraphics[width=\swidthfour]{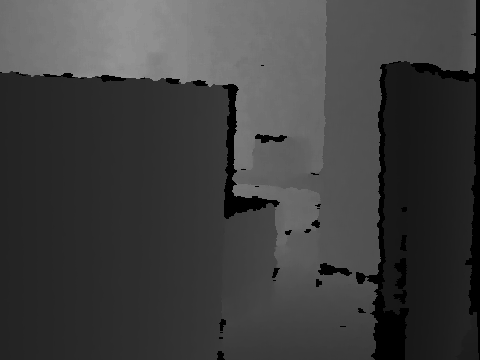}}\hspace{-6pt}
        \hfil \subfloat[GDF]{\label{fig.kinectdepth.blf}
        \includegraphics[width=\swidthfour]{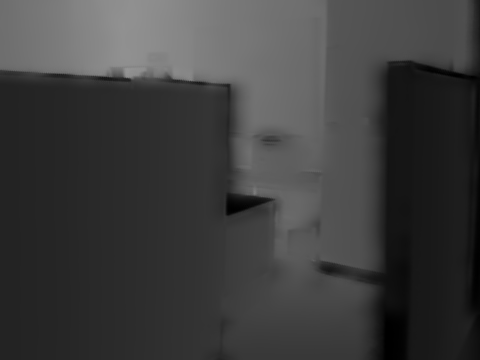}}\hspace{-6pt}
        \hfil \subfloat[Ours ({$\mathcal{F}$} is GDF)]{\label{fig.kinectdepth.gf}
        \includegraphics[width=\swidthfour]{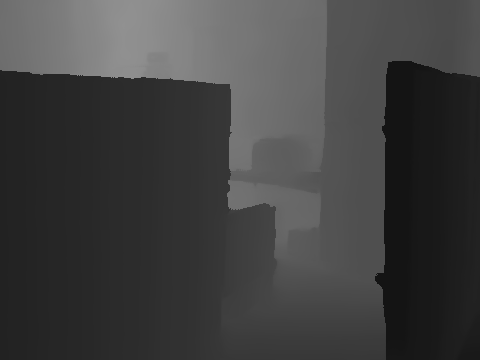}}}
    \caption{Joint filtering on depth map obtained by Microsoft Kinect camera. The parameters are $\sigma_s=10$, $\sigma_r=0.1$. The loss function in our framework is truncated $L_1$ norm. The invalid regions in the depth map (black holes) can be neatly fixed by our enhanced joint filtering (the value of invalid pixels is treated as zero during the filtering).}
    \label{fig.kinectdepth}
\end{figure*}

\subsection{Piecewise-constant Smoothing}\label{sec:specificblfgf}

When the operator $\mathcal{F}$ in Eq. \eqref{eq.ourframework} is bilateral filter or guided filter, the proposed framework is actually a \emph{weighted median filter} \cite{ma2013constant} (with $L_1$ norm) or a \emph{weighted mode filter} \cite{min2012depth} (with a redescending-influence loss function). The framework plays a role for ``enhancing'' the edge-preserving ability of $\mathcal{F}$ to achieve piecewise-constant smoothing, due to the derivation of the $M$-smoother from piecewise-constant model \cite{chu1998edge}. Fig. \ref{fig.noisesyn} shows such an example on a synthetic image. Fig. \ref{fig.enhancebfandgf} shows an example on a natural image. Note that although the loss function in our framework works like the range weighting function in bilateral filter, yet adding another range weighting kernel into bilateral filter (i.e., changing $\sigma_r$ in Gaussian range kernel) cannot yield our smoother. Also, performing bilateral filtering in an iterative manner cannot achieve the same results as our smoother (as discussed in Sec. \ref{sec:relatedworkrobust}). Compared with the histogram filters (e.g., when $\mathcal{F}$ is box filter or Gaussian filter in our framework), the new smoother can better preserve edges (recall that local histogram completely ignores the color/intensity value of center pixel). Fig. \ref{fig.compmodegf} gives an illustration.

\begin{figure*}[!t]
    \centerline{\subfloat[Original joint BLF (JBF) or guided filter (GDF)]{\label{fig.stereo_middlebury_numerical.sigr21}
    \includegraphics[width=\swidthtwo]{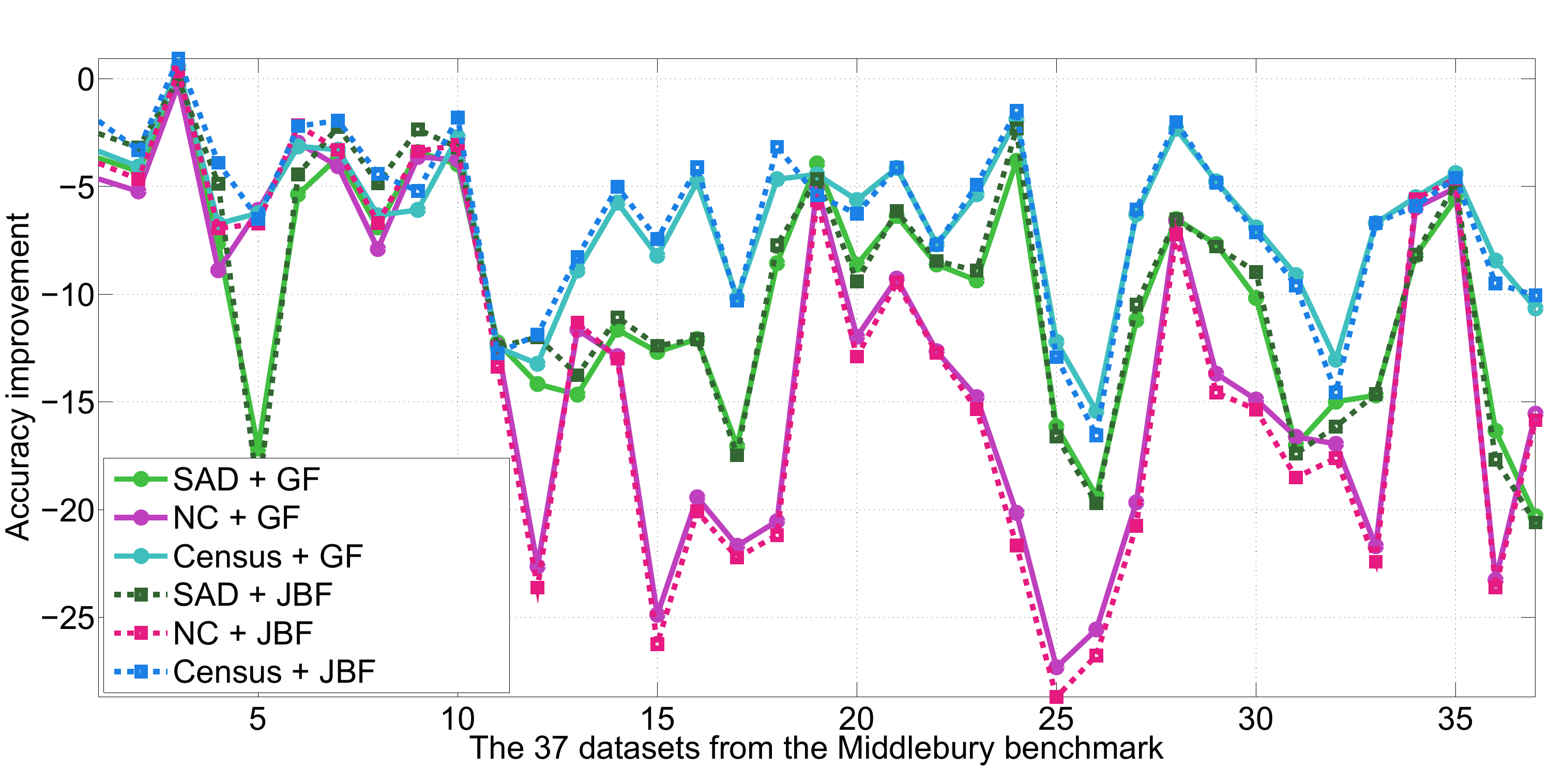}}
    \subfloat[Ours]{\label{fig.stereo_middlebury_numerical.sigr22}
    \includegraphics[width=\swidthtwo]{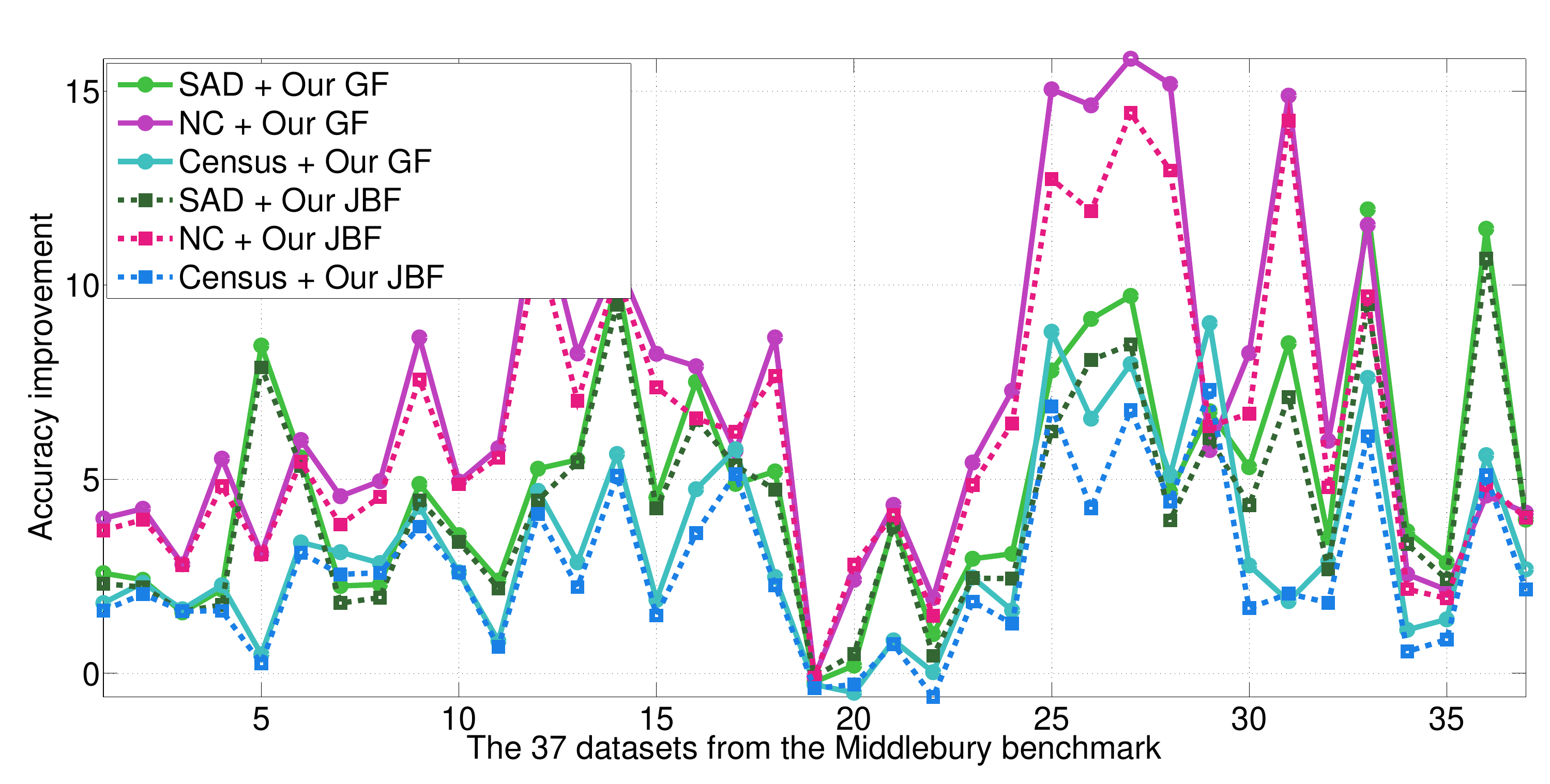}}}
    \caption{Quantitative comparison of disparity map refinement by original joint filtering and our enhanced joint filtering (accuracy is measured by percentage of bad pixels \cite{scharstein2002taxonomy}). The three stereo matching algorithms, i.e., Sum of Absolute Differences (SAD), Normalized Correlation (NC), Census Transform, can be computed very efficiently but the quality of the produced disparity map is low. Filtering the disparity map using joint bilateral filter or guided filter with input RGB image as guidance image, the quality of disparity map can not be improved. By contrast, our enhanced joint bilateral filtering or guided filtering (with truncated $L_1$ norm loss function) can commonly make the quality of the disparity map improved. An example of the visual comparison is provided in Fig. \protect\ref{fig:stereo_middlebury_visual}.}
    \label{fig.stereo_middlebury_numerical}
\end{figure*}

\begin{figure*}[!t]
    \centerline{\addtocounter{subfigure}{-1}\subfloat{
        \includegraphics[width=\swidthfive]{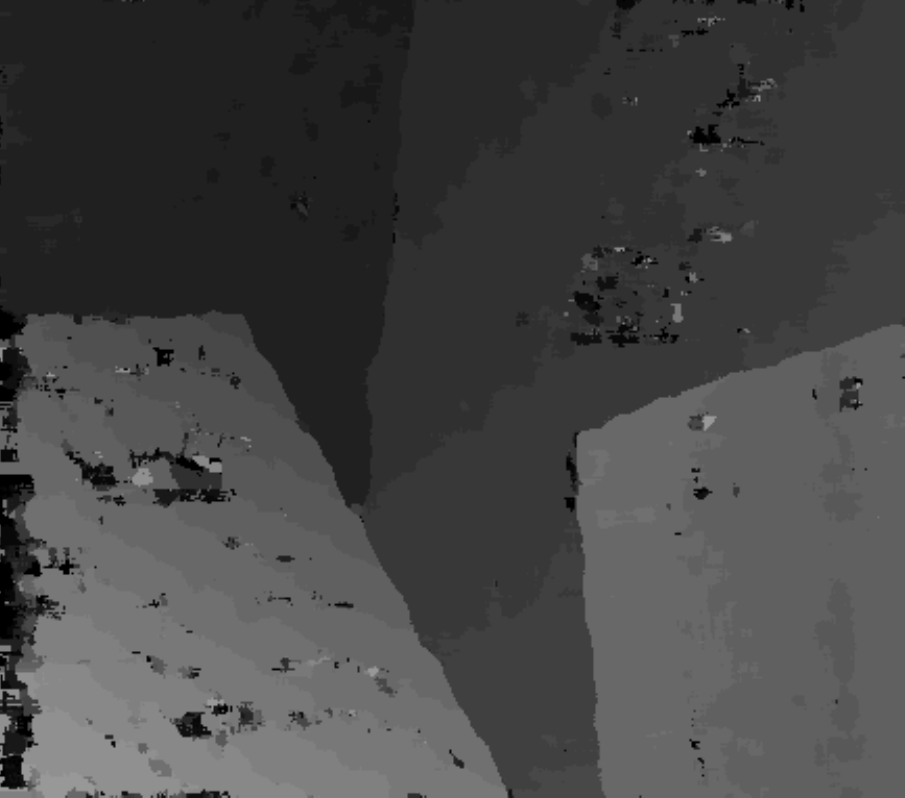}}\hspace{-15pt}
        \hfil \addtocounter{subfigure}{-1}\subfloat{
        \includegraphics[width=\swidthfive]{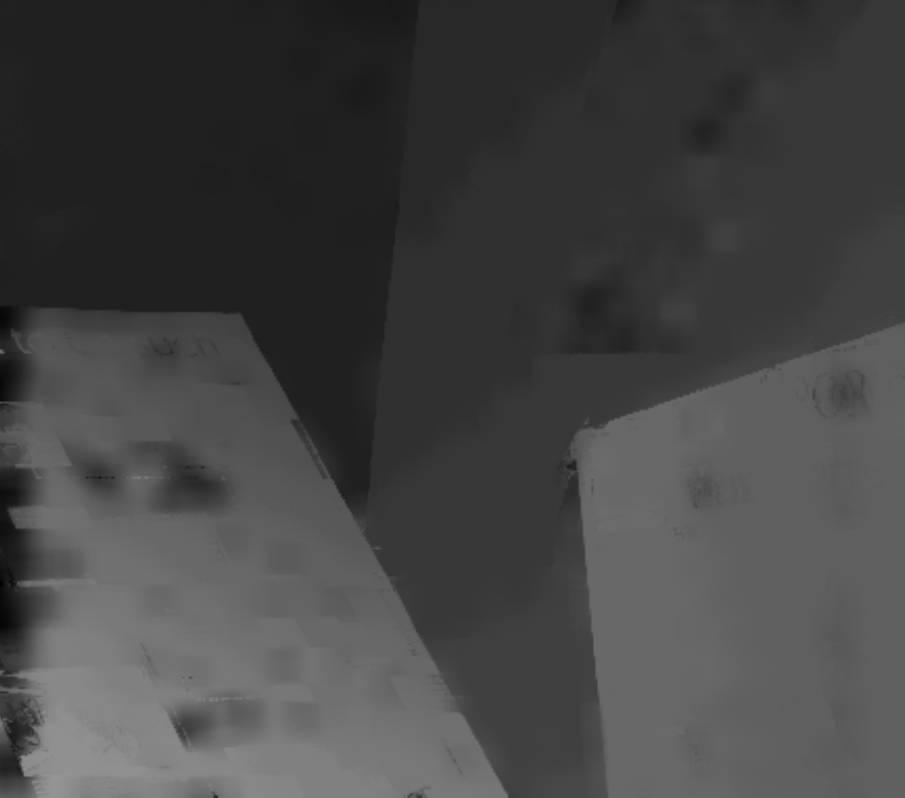}}\hspace{-15pt}
        \hfil \addtocounter{subfigure}{-1}\subfloat{
        \includegraphics[width=\swidthfive]{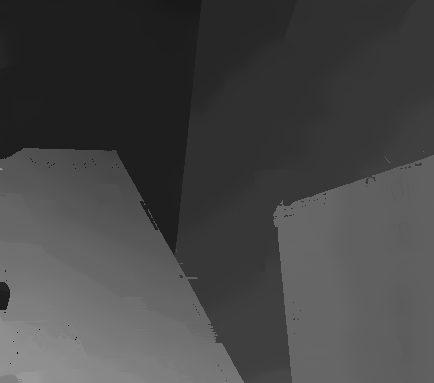}}\hspace{-15pt}
        \hfil \addtocounter{subfigure}{-1}\subfloat{
        \includegraphics[width=\swidthfive]{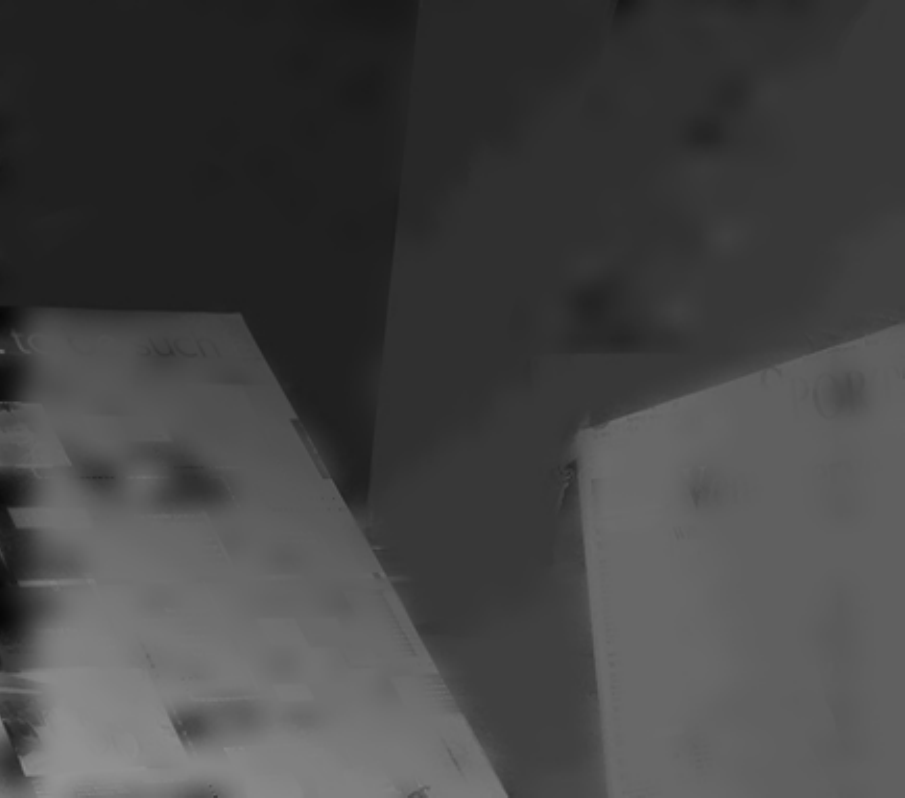}}\hspace{-15pt}
        \hfil \addtocounter{subfigure}{-1}\subfloat{
        \includegraphics[width=\swidthfive]{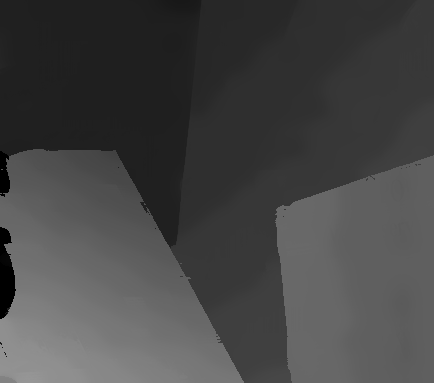}}}\vspace{-10pt}
    \centerline{\addtocounter{subfigure}{-1}\subfloat{
        \includegraphics[width=\swidthfive]{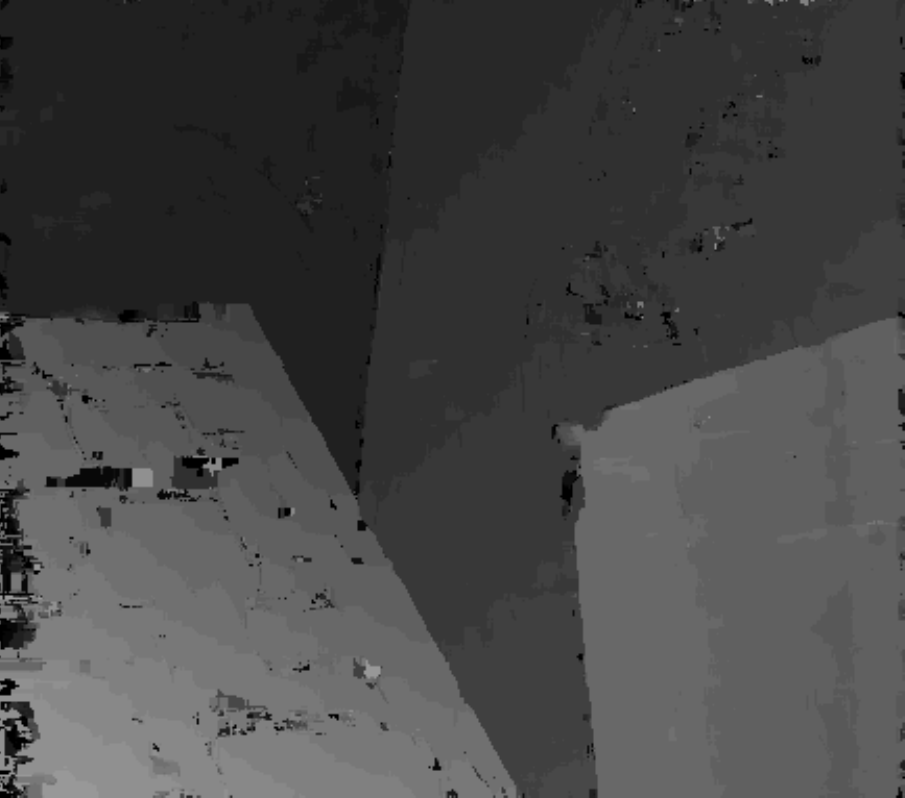}}\hspace{-15pt}
        \hfil \addtocounter{subfigure}{-1}\subfloat{
        \includegraphics[width=\swidthfive]{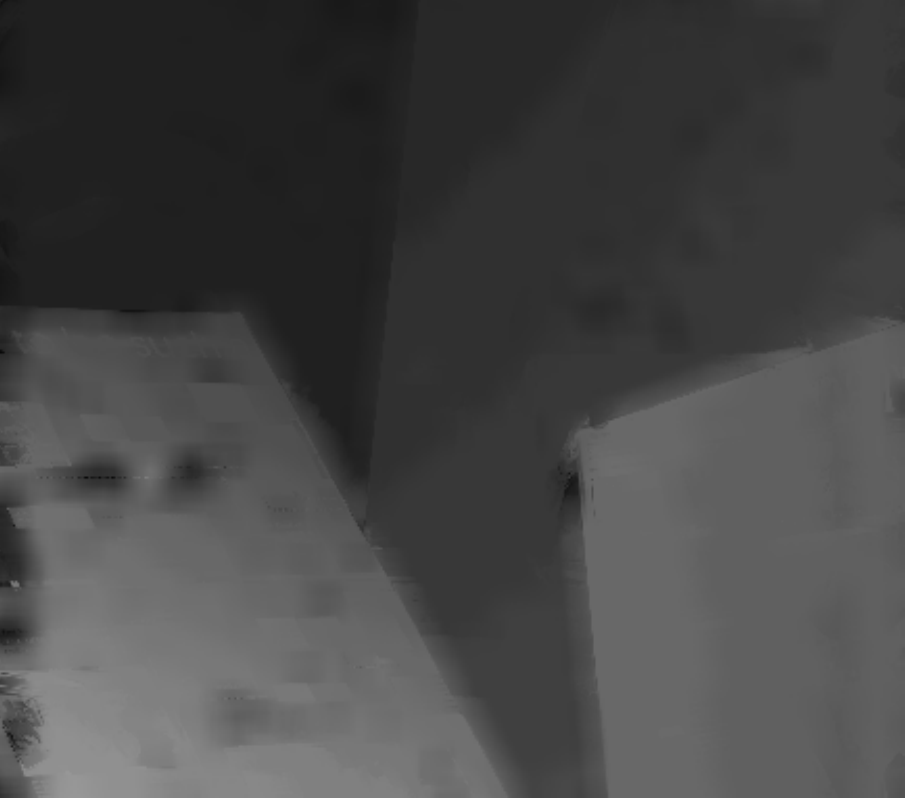}}\hspace{-15pt}
        \hfil \addtocounter{subfigure}{-1}\subfloat{
        \includegraphics[width=\swidthfive]{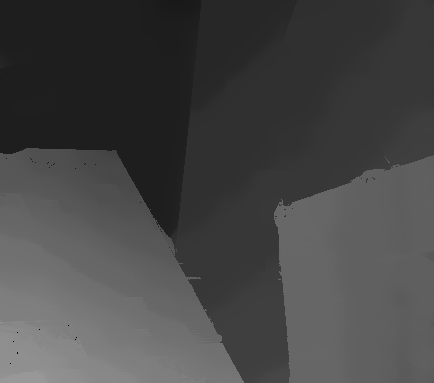}}\hspace{-15pt}
        \hfil \addtocounter{subfigure}{-1}\subfloat{
        \includegraphics[width=\swidthfive]{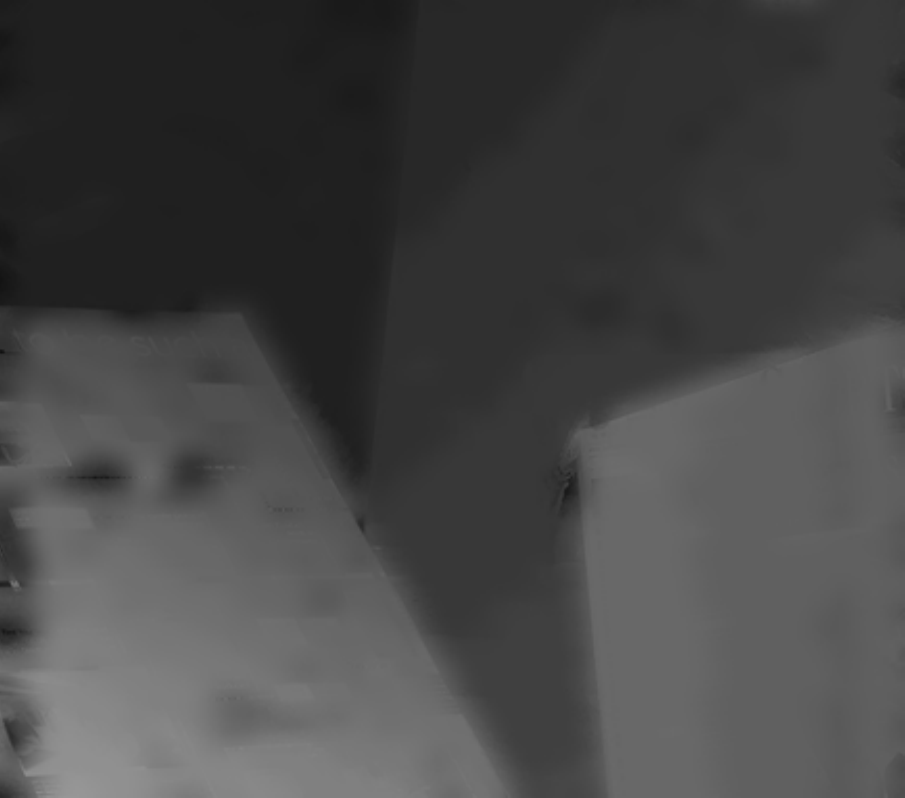}}\hspace{-15pt}
        \hfil \addtocounter{subfigure}{-1}\subfloat{
        \includegraphics[width=\swidthfive]{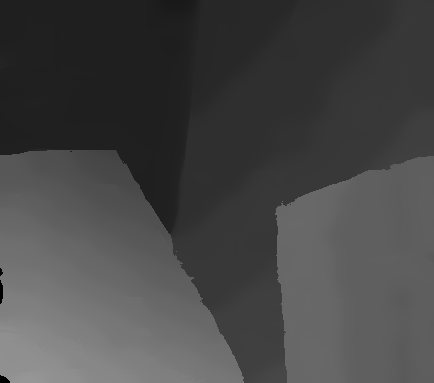}}}\vspace{-10pt}
    \centerline{\subfloat[Original disparity maps]{
        \includegraphics[width=\swidthfive]{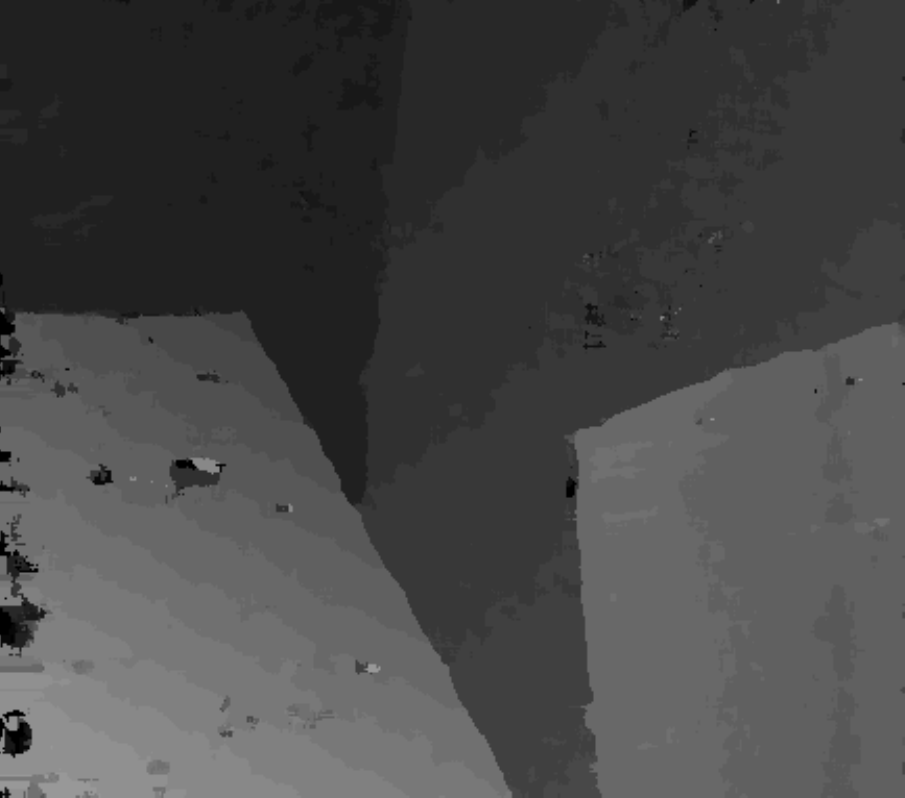}}\hspace{-15pt}
        \hfil \subfloat[JBF ($\sigma_s$=$10$,$\sigma_r$=$0.1$)]{
        \includegraphics[width=\swidthfive]{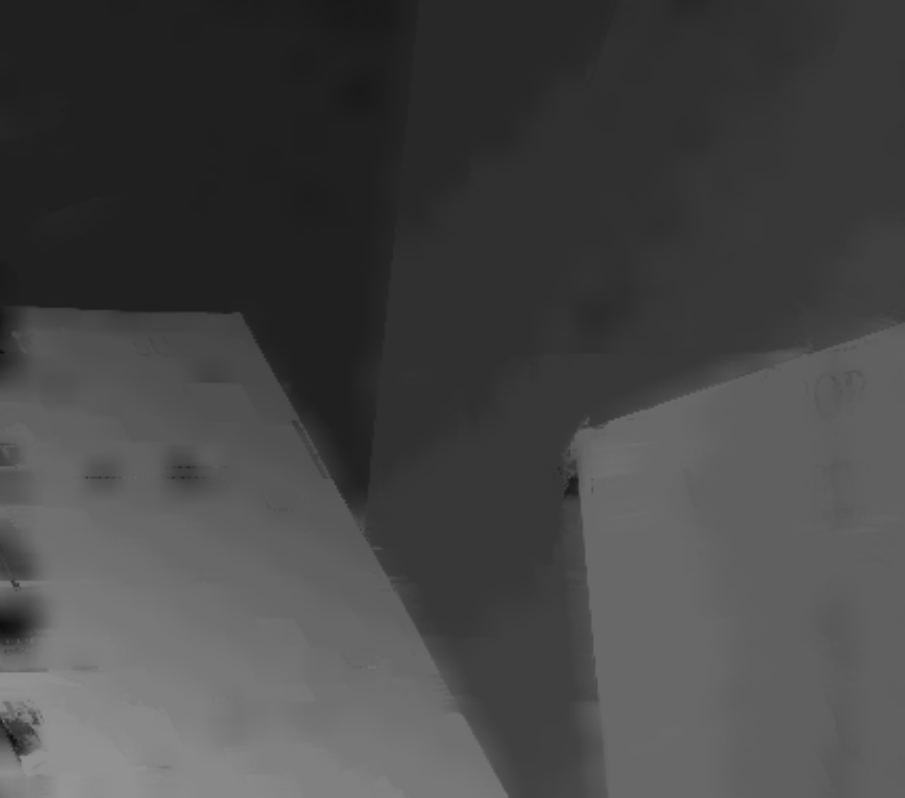}}\hspace{-15pt}
        \hfil \subfloat[Ours ($\mathcal{F}$ is JBF)]{
        \includegraphics[width=\swidthfive]{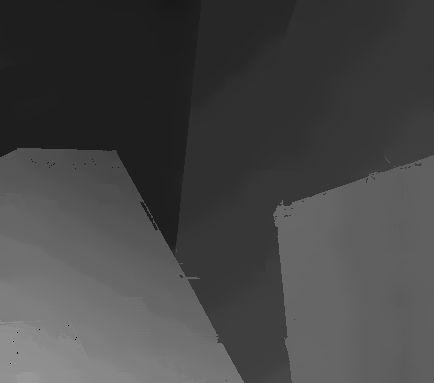}}\hspace{-15pt}
        \hfil \subfloat[GDF ($\sigma_s$=$10$,$\sigma_r$=$0.1$)]{
        \includegraphics[width=\swidthfive]{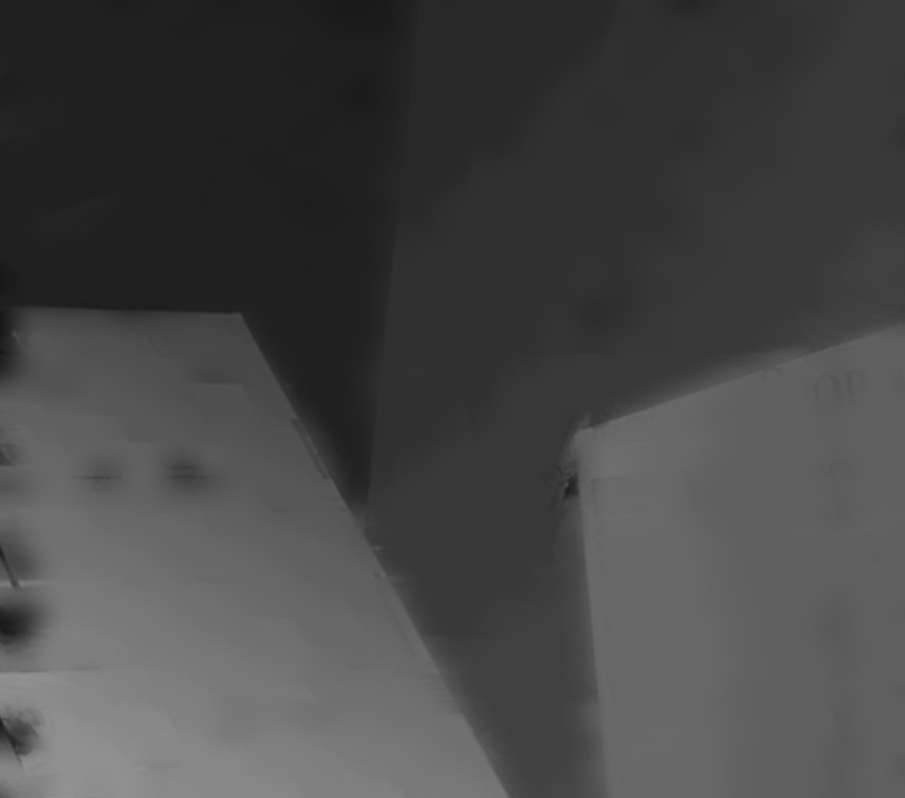}}\hspace{-15pt}
        \hfil \subfloat[Ours ($\mathcal{F}$ is GDF)]{
        \includegraphics[width=\swidthfive]{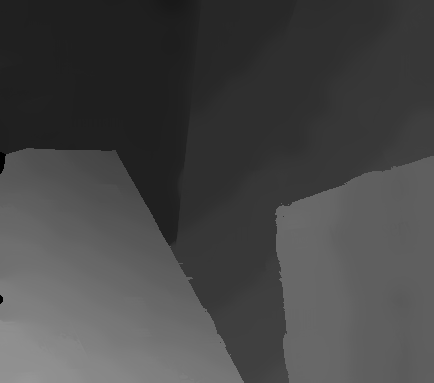}}}
	\caption{Visual comparison of disparity map refinement for \textit{Venus} dataset from Middlebury benchmark \cite{scharstein2002taxonomy}. The three rows correspond to the three algorithms in Fig. \ref{fig.stereo_middlebury_numerical}. Loss function in our framework is truncated $L_1$ norm.}
    \label{fig:stereo_middlebury_visual}
\end{figure*}

Additionally, both the (joint) bilateral filtering and guided filtering can use another image \cite{min2012depth}, rather than the input image itself, as guidance image. This gives the filtering more flexibility (hereafter referred to as \emph{joint filtering}) and makes it especially suitable for depth image restoration. For example, the enhanced joint filtering can be employed to reduce the noise of depth image, which can be acquired by commercial cameras but is commonly noisy, using clean RGB image that can be simultaneously acquired with depth image as guidance image. Fig. \ref{fig.depthdenoising} shows two experimental results of the joint bilateral filtering on disparity maps (disparity is inversely proportional to depth). The numerical comparison shows the effectiveness of our approach. Fig. \ref{fig.kinectdepth} shows a real world example of the denoising on depth image obtained by Microsoft Kinect camera. Note that the invalid regions in the original depth map (black holes) can be neatly fixed by our approach. Similarly, the enhanced joint filtering can also serve as a tool for refining disparity maps produced by existing stereo matching algorithms \cite{scharstein2002taxonomy}. Fig. \ref{fig.stereo_middlebury_numerical} shows a quantitative comparison of the disparity map refinement on the Middlebury benchmark \cite{scharstein2002taxonomy} (we use three basic and efficient stereo matching algorithms to produce low-quality disparity maps), using original joint filtering and our enhanced joint filtering, respectively. Compared to the original joint filtering, our approach can produce cleaner and sharper disparity maps. With a CUDA implementation, our enhanced joint filtering can achieve real-time performance on GPU (e.g., the enhanced guided filtering with $n=16$ takes about $30$ ms/Mp on our GPU).

\section{Concluding Remarks}\label{sec:conclusion}

We have presented a filtering framework for achieve piecewise-constant smoothing. The proposed framework is derived by solving a generalized $M$-smoother and can be implemented very efficiently on modern many-core processors utilizing parallelism. We demonstrate the effectiveness of the proposed framework for fast approximation of local-histogram-based filters and enhancing existing edge-preserving filters. An unsolved problem is how to further reduce the number of filtering required in the framework, for example, by automatically selecting the sampling intensity levels to minimize the quantization error. We intend to investigate this problem in the future.

\appendix[Derivation of Parabolic Fitting Approximation]

Let us first consider the simplest case Eq. \eqref{eq.l1estimate} (box filter with $L_1$ norm loss function). Since neighboring pixels tend to be similar to each other, we assume the pixel values near $\mathbf{p}$ follow a uniform distribution between $a$ and $b$ ($0 < a < b < 255$). Then the cost function of Eq. \eqref{eq.l1estimate} can be rewritten as an integral
\begin{eqnarray*}
    E(\theta) &=& \sum\limits_{\mathbf{q}\in\Omega_\mathbf{p}} |\theta-I_\mathbf{q}| \approx \int_{a}^{b} |\theta-x| \mathrm{d} x \\
    &=& \left\{ \begin{matrix} \frac{1}{2}[ (\theta - b)^2 - (\theta - a)^2)], \text{~if~} \theta < a \\ \frac{1}{2}[ (\theta - a)^2 + (\theta - b)^2)], \text{~if~} a \leqslant \theta \leqslant b \\ \frac{1}{2}[ (\theta - a)^2 - (\theta - b)^2)], \text{~if~} \theta > b \end{matrix} \right.
\end{eqnarray*}
The function is a ``cup-shaped'' continuous function, with two linear segments in the ends and a quadratic segment in the middle. Thus in our approximate algorithm, we can fit a parabola near the bottom of the ``cup-shaped'' function to find the minima. The derivation can be generalized to other loss functions in Table \ref{tab.lossfuncshow} (although their resulting cost function will not be parabolic curve, in our algorithm we use parabolic fitting to approximate all the cases for simplicity).

Notice that the above derivation is based on a much simplified problem (uniform distributed pixel values near $\mathbf{p}$). In fact, because the cost function is data-dependent (depending on the neighboring pixels near $\mathbf{p}$), the actual cost function cannot be analytically solved. Besides, the complicated weighting schemes other than box filter (see Table \ref{tab:wafall}) will add more complexity to the problem. Thus we conducted a thorough experimental validation in Section \ref{sec:experiments}. The experiments show that the approximate algorithm works well in practice.



\bibliographystyle{abbrv}
\bibliography{TCSVT_double_rev3_arXiv}

\begin{thebibliography}{10}

\bibitem{barash2002fundamental}
D.~Barash.
\newblock Fundamental relationship between bilateral filtering, adaptive
  smoothing, and the nonlinear diffusion equation.
\newblock {\em {IEEE} Trans. Pattern Anal. Mach. Intell.}, 24(6):844--847,
  2002.

\bibitem{black1996unification}
M.~Black and A.~Rangarajan.
\newblock On the unification of line processes, outlier rejection, and robust
  statistics with applications in early vision.
\newblock {\em Int. J. Comput. Vis.}, 19(1):57--91, 1996.

\bibitem{black1998robust}
M.~Black, G.~Sapiro, D.~Marimont, and D.~Heeger.
\newblock Robust anisotropic diffusion.
\newblock {\em {IEEE} Trans. Image Process.}, 7(3):421--432, 1998.

\bibitem{Chaudhury2011blf}
K.~Chaudhury, D.~Sage, and M.~Unser.
\newblock Fast o(1) bilateral filtering using trigonometric range kernels.
\newblock {\em {IEEE} Trans. Image Process.}, 20(12):3376--3382, Dec 2011.

\bibitem{chu1998edge}
C.~Chu, I.~Glad, F.~Godtliebsen, and J.~Marron.
\newblock Edge-preserving smoothers for image processing.
\newblock {\em Journal of the American Statistical Association},
  93(442):526--541, 1998.

\bibitem{crow1984summed}
F.~Crow.
\newblock Summed-area tables for texture mapping.
\newblock {\em ACM Trans. Graph. (Proc. SIGGRAPH)}, 18(3):207--212, 1984.

\bibitem{deriche1992recursively}
R.~Deriche.
\newblock Recursively implementing the gaussian and its derivatives.
\newblock In {\em ICIP 1992}, pages 263--267, 1992.

\bibitem{durand2002fast}
F.~Durand and J.~Dorsey.
\newblock Fast bilateral filtering for the display of high-dynamic-range
  images.
\newblock {\em ACM Trans. Graph. (Proc. SIGGRAPH)}, 21(3):257--266, 2002.

\bibitem{elad2002origin}
M.~Elad.
\newblock On the origin of the bilateral filter and ways to improve it.
\newblock {\em {IEEE} Trans. Image Process.}, 11(10):1141--1151, 2002.

\bibitem{hampel1986robust}
F.~Hampel, E.~Ronchetti, P.~Rousseeuw, and W.~Stahel.
\newblock {\em Robust statistics: the approach based on influence functions}.
\newblock Wiley, 1986.

\bibitem{he2010guided}
K.~He, J.~Sun, and X.~Tang.
\newblock Guided image filtering.
\newblock In {\em ECCV}, pages 1--14, 2010.

\bibitem{he2013guidedpami}
K.~He, J.~Sun, and X.~Tang.
\newblock Guided image filtering.
\newblock {\em {IEEE} Trans. Pattern Anal. Mach. Intell.}, 35(6):1397--1409,
  2013.

\bibitem{huber1981robust}
P.~Huber.
\newblock {\em Robust statistics}.
\newblock Wiley, 1981.

\bibitem{jiang14mixednoise}
J.~Jiang, L.~Zhang, and J.~Yang.
\newblock Mixed noise removal by weighted encoding with sparse nonlocal
  regularization.
\newblock {\em {IEEE} Trans. Image Process.}, 23(6):2651--2662, June 2014.

\bibitem{kass2010smoothed}
M.~Kass and J.~Solomon.
\newblock Smoothed local histogram filters.
\newblock {\em ACM Trans. Graph. (Proc. SIGGRAPH)}, 29(4):100:1--100:10, 2010.

\bibitem{lu2012cross}
J.~Lu, K.~Shi, D.~Min, L.~Lin, and M.~N. Do.
\newblock Cross-based local multipoint filtering.
\newblock In {\em CVPR}, pages 430--437. IEEE, 2012.

\bibitem{ma2013constant}
Z.~Ma, K.~He, Y.~Wei, J.~Sun, and E.~Wu.
\newblock Constant time weighted median filtering for stereo matching and
  beyond.
\newblock In {\em ICCV}, pages 49--56. IEEE, 2013.

\bibitem{min2012depth}
D.~Min, J.~Lu, and M.~N. Do.
\newblock Depth video enhancement based on weighted mode filtering.
\newblock {\em {IEEE} Trans. Image Process.}, 21(3):1176--1190, 2012.

\bibitem{mrazek2006robust}
P.~Mr{\'a}zek, J.~Weickert, and A.~Bruhn.
\newblock On robust estimation and smoothing with spatial and tonal kernels.
\newblock {\em Geometric Properties for Incomplete Data}, pages 335--352, 2006.

\bibitem{parisblfdataurl}
S.~Paris.
\newblock \url{http://people.csail.mit.edu/sparis/bf/#code}, 2013.
\newblock [Online; accessed 2-Feb-2013].

\bibitem{paris2009fast}
S.~Paris and F.~Durand.
\newblock A fast approximation of the bilateral filter using a signal
  processing approach.
\newblock {\em Int. J. Comput. Vis.}, 81(1):24--52, 2009.

\bibitem{paris2009bilateralbook}
S.~Paris, P.~Kornprobst, and J.~Tumblin.
\newblock {\em Bilateral filtering: Theory and applications}, volume~1.
\newblock Now Publishers Inc, 2009.

\bibitem{perona1990scale}
P.~Perona and J.~Malik.
\newblock Scale-space and edge detection using anisotropic diffusion.
\newblock {\em {IEEE} Trans. Pattern Anal. Mach. Intell.}, 12(7):629--639,
  1990.

\bibitem{rhemann2011fast}
C.~Rhemann, A.~Hosni, M.~Bleyer, C.~Rother, and M.~Gelautz.
\newblock Fast cost-volume filtering for visual correspondence and beyond.
\newblock In {\em CVPR 2011}, pages 3017--3024. IEEE, 2011.

\bibitem{scharstein2002taxonomy}
D.~Scharstein and R.~Szeliski.
\newblock A taxonomy and evaluation of dense two-frame stereo correspondence
  algorithms.
\newblock {\em Int. J. Comput. Vis.}, 47(1):7--42, 2002.

\bibitem{tomasi1998bilateral}
C.~Tomasi and R.~Manduchi.
\newblock Bilateral filtering for gray and color images.
\newblock In {\em ICCV 1998}, pages 839--846. IEEE, 1998.

\bibitem{van2001local}
J.~Van~de Weijer and R.~Van~den Boomgaard.
\newblock Local mode filtering.
\newblock In {\em CVPR 2001}, volume~2, pages II--428. IEEE, 2001.

\bibitem{winkler1999noise}
G.~Winkler, V.~Aurich, K.~Hahn, A.~Martin, and K.~Rodenacker.
\newblock Noise reduction in images: Some recent edge-preserving methods.
\newblock {\em Pattern Recognition and Image Analysis}, 9(4):749--766, 1999.

\bibitem{yang2009real}
Q.~Yang, K.~Tan, and N.~Ahuja.
\newblock Real-time o (1) bilateral filtering.
\newblock In {\em CVPR 2009}, pages 557--564. IEEE, 2009.

\bibitem{cvpr-07-qingxiong-yang}
Q.~Yang, R.~Yang, J.~Davis, and D.~Nist{\'e}r.
\newblock Spatial-depth super resolution for range images.
\newblock In {\em CVPR 2007}. IEEE, 2007.

\bibitem{zuo2014gradienttip}
W.~Zuo, L.~Zhang, C.~Song, D.~Zhang, and H.~Gao.
\newblock Gradient histogram estimation and preservation for texture enhanced
  image denoising.
\newblock {\em {IEEE} Trans. Image Process.}, 23(6):2459--2472, June 2014.

\end{thebibliography}

%

\end{document}